%% file: apt.tex
\documentclass{article}

\usepackage[final]{corl_2026} 
\usepackage{enumerate}
\usepackage{algorithmic}
\usepackage{algorithm}
\usepackage{array}
\usepackage{textcomp}
\usepackage{stfloats}
\usepackage{natbib}
\usepackage{xspace}
\usepackage{url}
\usepackage{color}
\usepackage{subfigure}
\usepackage{makecell}
\usepackage{multirow}
\usepackage{verbatim}
\usepackage{graphicx}
\usepackage{mathrsfs}
\usepackage{bbding}
\usepackage{enumitem}
\usepackage{amsmath,amssymb,amsthm}
\usepackage{gensymb}
\usepackage{dsfont}
\usepackage{threeparttable}
\usepackage{booktabs}
\usepackage{colortbl}
\usepackage{wrapfig}
\usepackage[table,xcdraw]{xcolor}
\usepackage{url}
\usepackage{caption}

 \newcommand{\ours}{APT\xspace}  

\title{APT: Action Expert Pretraining Improves Instruction Generalization of Vision-Language-Action Policies}

\author{
  {Kechun Xu$^{1,2}$, Zhenjie Zhu$^1$, Anzhe Chen$^1$, Rong Xiong$^{1,3}$, Yue Wang$^1$}\\
  $^1$Zhejiang University, $^2$RobbyAnt, $^3$Zhejiang Humanoid Robot Innovation Center \\[6pt]
  \url{https://xukechun.github.io/papers/APT}
}

\begin{document}
\maketitle

\input{sec/abstract}
\input{sec/introduction}
\input{sec/related}

\input{sec/method}
\input{sec/experiment}
\input{sec/conclusion}

\input{sec/acknowledgment}

\bibliography{ref}  

\input{sec/appendix}

\end{document}

%% file: sec/abstract.tex
\begin{abstract}
Vision-Language-Action~(VLA) models that couple pretrained Vision-Language Models~(VLMs) with continuous action experts have achieved strong manipulation performance, yet generalization to out-of-distribution~(OOD) language instructions remains poor. A known challenge is the structural imbalance in VLA data, where language is far less diverse than visual and action content, making policies prone to visual shortcuts. While discrete-action methods mitigate this through vision-language co-training, continuous action experts lack such protection: they start from random initialization and learn entirely from imbalanced data, producing noisy gradients that corrupt the VLM and fail to exploit its language capability. We address this from a Bayesian perspective, factorizing the policy into a language-agnostic Vision-Action~(VA) prior and a language-conditioned VLA likelihood, and propose \textbf{APT}, a two-stage training method emphasizing \textbf{A}ction expert \textbf{P}re\textbf{T}raining. In Stage~1, the action expert is pretrained as a VA prior on vision-action pairs from a frozen VLM, bypassing the language imbalance. In Stage~2, language tokens are injected through a gated fusion mechanism that integrates VLM features while preserving the learned visuomotor prior. APT applies to mainstream VLA architectures, including the $\pi$ and GR00T-style architectures. Comprehensive experiments validate that APT achieves consistent gains on unseen instructions and compositional tasks.

\end{abstract}

\keywords{VLA, Language Generalization, Manipulation} 

%% file: sec/introduction.tex
\section{Introduction}

\begin{wrapfigure}{r}{7cm}
\centering
\vspace{-0.5cm}
\includegraphics[width=0.5\textwidth]{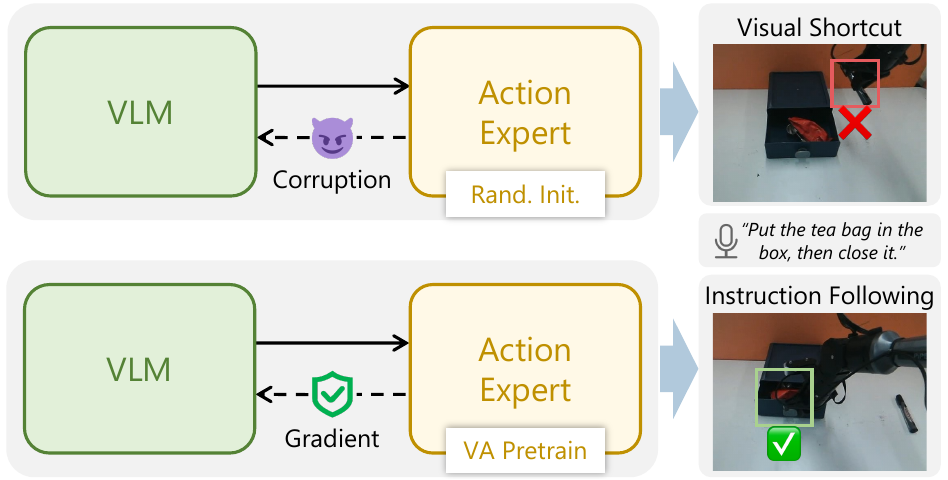}
\vspace{-0.45cm}
\caption{Action expert pretraining~(APT) enables effective instruction following.}
\vspace{-0.5cm}
\label{fig:teaser}
\end{wrapfigure}

Enabling robots to follow diverse task instructions across varied environments is a long-standing goal of generalizable robot policies. Vision-Language-Action~(VLA) models have emerged as a promising paradigm toward this goal, leveraging pretrained Vision-Language Models~(VLMs) to ground language instructions in visual observations and generate executable robot actions~\cite{black2024pi0,bjorck2025gr00tn1,generalist2025gen0}. A critical requirement for real-world deployment is out-of-distribution~(OOD) language generalization, \textit{i.e.}, reliably following instructions and task compositions unseen during training. Despite significant recent progress, this capability remains largely unresolved~\cite{zhou2025liberopro,gao2025taxonomy}: policies often fail under unseen paraphrased commands, novel object references, or compositional task specifications.

To achieve language generalization, a prerequisite is that the model genuinely attends to language instructions. However, prior work has identified a structural \textit{imbalance} in VLA data~\cite{xu2025bayesvla,fang2026vision}: most trajectories pair many vision-action frames with a single task instruction, making policies prone to visual shortcuts that bypass language. The discrete-action paradigm, which extends a pretrained VLM with discretized action tokens~\cite{zitkovich2023rt2,kim2024openvla,pertsch2025fast,team2025geminirobotics,zhao2025cotvla}, mitigates this through VL co-training on large-scale reasoning data, effectively preserving the VLM's language capability, and can demonstrate good instruction following~\cite{pertsch2025fast}. However, discrete representations struggle with the continuous, dynamic nature of dexterous manipulation. The continuous-action paradigm addresses this by coupling the VLM with a generative action expert~\cite{black2024pi0,bjorck2025gr00tn1,liu2024rdt,wen2025diffusionvla}, substantially improving action quality. However, the action expert lacks both pretraining and availability of co-training data, and must learn entirely from imbalanced VLA data. Recent work further shows that gradients from this randomly initialized action expert can be harmful to the VLM backbone, and proposes stopping such gradients~\cite{black2025pi_ohfive,driess2025knowledge,bjorck2025gr00tn15}. We explain that this gradient harm originates from training a randomly initialized action expert on imbalanced VLA data, where it gravitates toward visual shortcuts, thus producing noisy gradients that corrupt the VLM~(Figure~\ref{fig:teaser}). By analogy with the discrete-action paradigm, where VL pretraining protects against such shortcuts, we hypothesize that properly pretraining the action expert can relieve this shortcut tendency and improve instruction following. This raises a question: \textit{how can we pretrain the action expert using only the VLA data?}

We answer this through a Bayesian factorization of the VLA policy, $\pi(\mathbf{a} | \mathbf{v}, \ell) \propto \pi^p(\mathbf{a} | \mathbf{v}) \cdot \mathcal{L}(\ell | \mathbf{v}, \mathbf{a})$, which separates action generation into a language-agnostic Vision-Action~(VA) prior $\pi^p(\mathbf{a}|\mathbf{v})$ and a language-conditioned VLA likelihood $\mathcal{L}(\ell | \mathbf{v}, \mathbf{a})$. The key observation is that although full VLA triplets suffer from language-vision imbalance, vision-action pairs alone are \textit{well-balanced} and do not create shortcut incentives. Pretraining the action expert as a VA prior on vision-action pairs therefore provides a principled initialization, before any language is introduced. Building on this prior, the VLA likelihood is obtained by finetuning the action expert through VLM language token injection, starting from a point where the action expert already captures coherent visuomotor control. 

We propose \textbf{\ours}, a simple yet effective two-stage training method, emphasizing \textbf{A}ction expert \textbf{P}re-\textbf{T}raining. As shown in Figure~\ref{fig:overview}, in Stage~1, a diffusion-based action expert is pretrained as a VA prior conditioned solely on visual tokens from a frozen VLM backbone. In Stage~2, language tokens are introduced alongside additional attention layers, and the full policy is jointly finetuned to condition action generation on task instructions, implicitly forming the VLA likelihood. We propose a novel action expert design, characterized by a gated fusion mechanism that integrates layer-wise VLM features into the action expert, enabling the model to inherit the VLM's representational capacity while preserving the pretrained visuomotor priors. We find that jointly finetuning action expert with VLM from a pretrained VA prior yields substantially better language generalization than training from random action initialization, confirming the effectiveness of action expert pretraining. Notably, the two-stage training applies to mainstream continuous-action VLA architectures, such as the $\pi$~\cite{black2024pi0,black2025pi_ohfive} and GR00T~\cite{bjorck2025gr00tn1,bjorck2025gr00tn15}-style architectures, consistently improving their language generalization. Extensive experiments across simulation and real-world settings validate improved generalization to unseen instructions and unseen compositional tasks. To summarize, the main contributions are:

\begin{itemize}[leftmargin=*, topsep=2pt, itemsep=1pt]
    \item We propose APT, a two-stage training method motivated by a Bayesian perspective, enabling action expert pretraining on vision-action pairs within existing VLA datasets.
    \item We propose a novel action expert design with a layer-wise gated fusion mechanism for effective VLM feature fusion.
    \item We demonstrate that action expert pretraining generalizes across architectures and achieves consistent gains on OOD language generalization in extensive simulation and real-world experiments.
\end{itemize}

%% file: sec/related.tex
\section{Related Work}
\label{sec:related}
\textbf{Vision-Language-Action Models.} Vision-Language-Action~(VLA) models have emerged as a dominant paradigm for generalist robot policies~\cite{zhong2025surveyvla,liu2025robovlms,cui2025openhelix}. Early works~\cite{brohan2022rt1,team2024octo,zhao2023act,jiang2023vima} train transformer-based policies from scratch on large-scale robot datasets~\cite{openxembodiment,khazatsky2024droid,fang2024rh20t,wu2024robomind,bu2025agibotworld,chen2025robotwin2,geng2025roboverse}. Subsequent advances finetune pretrained VLMs~\cite{qwen3vl2025,beyer2024paligemma} with action discretization~\cite{zitkovich2023rt2,kim2024openvla,zhao2025cotvla,kim2025oft}, retaining language capability through VL co-training~\cite{pertsch2025fast}, but discrete representations struggle with continuous dexterous manipulation. To improve action quality, diffusion- or flow-based continuous action experts have been widely adopted~\cite{chi2023dp,black2024pi0,liu2024rdt,zhu2025scaledp,liu2025hybridvla}, and dual-system architectures that combine VLM and continuous motor control become prevalent~\cite{black2024pi0,bjorck2025gr00tn1,bu2024robodual,shukor2025smolvla,wen2025diffusionvla,generalist2025gen0,jiang2025rynnvla}. Recent advances further scale this paradigm through VLM reasoning co-training~\cite{black2025pi_ohfive,bjorck2025gr00tn15,driess2025knowledge,wu2026pragmatic} and world-model auxiliary objectives~\cite{cen2025worldvla,zhang2025dreamvla,sun2026vlajepa,intelligence2026pi07,li2026causal}. Despite strong in-distribution performance, these models struggle to generalize to unseen instructions, objects, and compositional task variations~\cite{liu2023libero,zhou2025liberopro,fei2025liberoplus,nasiriany2024robocasa,gao2025taxonomy,guo2025robustness}. This is because the action experts are randomly initialized and co-trained with the VLM backbone, distorting pretrained language representations. This paper identifies this initialization gap as an important reason for language generalization failure and addresses it directly.

\textbf{Generalization on Task Instructions.} Benchmarking studies~\cite{zhou2025liberopro,orjuela2026brittle,gao2025taxonomy,wanna2025lang,fang2025intention} reveal that most VLA models treat language instructions as task identifiers rather than grounded semantic descriptions: high success on seen instructions collapses on novel objects, skills, or compositional variations. This reflects a structural shortcut where joint training exploits visual correlations while discarding language semantics~\cite{xu2025bayesvla,lian2026langforce}. A typical line to address this is knowledge insulation: stopping gradients from the
action expert to the VLM backbone so that robot finetuning does not corrupt pretrained language representations~\cite{driess2025knowledge,bjorck2025gr00tn15}. A complementary remedy is co-training on VL reasoning corpora to further regularize the VLM backbone~\cite{black2025pi_ohfive,yang2025instructvla,cheang2025gr3}. Architectural approaches such as text-aware feature extraction~\cite{huangotter} and inference-time language steering~\cite{nakamoto2024steering,wu2025foresight,zhang2025align,fang2026vision,zhan2026stable} offer
complementary gains. BayesVLA~\cite{xu2025bayesvla} provides a principled diagnosis via Bayesian factorization, separating a vision-action prior from a language-conditioned likelihood to prevent visual shortcut learning, and demonstrates generalization to novel instructions, but its pre-/post-contact decomposition limits scalability to heterogeneous datasets. LangForce~\cite{lian2026langforce} maximizes conditional mutual information between actions and instructions but focuses on seen-task following, paying less attention on OOD task generalization. This paper extends the Bayesian framework to an action pretraining perspective, identifying action expert initialization as a key factor for OOD language generalization, and addressing it through large-scale pretraining on balanced vision-action pairs.

%% file: sec/method.tex
\section{Method}
\label{sec:method}

\begin{figure}[t]
  \centering
  \includegraphics[width=\textwidth]{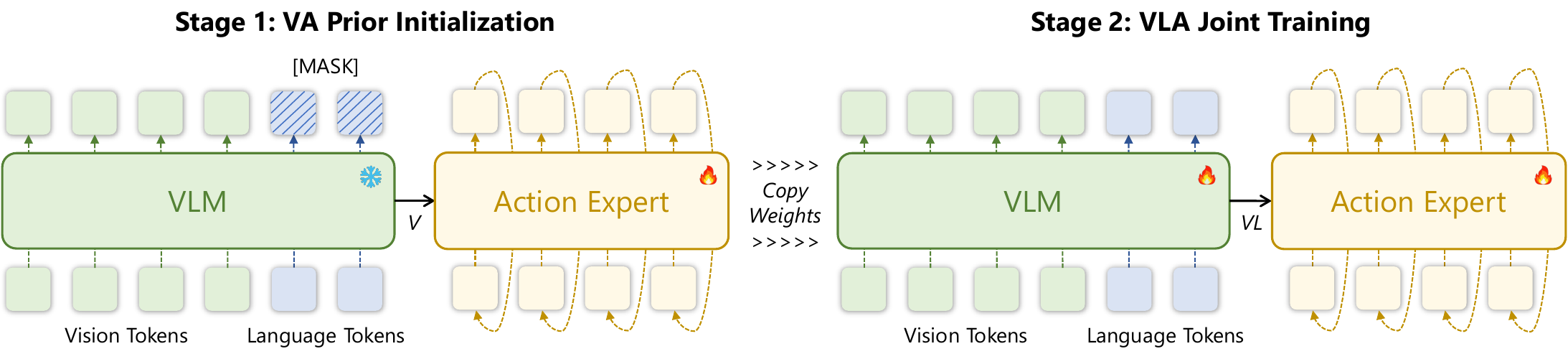}
  \vspace{-0.5cm}
  \caption{ Overview of APT. In Stage 1, the action expert is pretrained as a VA prior conditioned solely on visual tokens from a frozen VLM backbone. In Stage 2, language tokens are injected, training the full VLA policy to align the pretrained action distribution with the task instruction. 
  }
  \label{fig:overview}
\vspace{-0.5cm}
\end{figure}

\subsection{Problem Statement}
\label{sec:formulation}

We formulate the VLA policy as $\pi(\mathbf{a} \mid \mathbf{v}, \ell)$, where $\mathbf{a}$ is the robot action, $\mathbf{v}$ is the visual observations, and $\ell$ is the language instruction. We denote a dataset $D_{\mathrm{VLA}}$ that contains triplets $(\mathbf{a}, \mathbf{v}, \ell)$ for training the VLA policy. The standard VLA training objective is:
\begin{equation}
  \min\; -\mathbb{E}_{(\mathbf{a},\mathbf{v},\ell) \sim \mathcal{D}_{\mathrm{VLA}}}
  \bigl[\log \pi(\mathbf{a} \mid \mathbf{v}, \ell)\bigr].
  \label{eq:vla_obj}
\end{equation}
where $\mathbb{E}$ is the expectation on the dataset $D_\mathrm{VLA}$. In the dominant paradigm of continuous action generation~\cite{black2024pi0, bjorck2025gr00tn1}, a large pretrained VLM encodes visual and language tokens, which an action expert then consumes to generate continuous robot actions. Typically, there is no pretrained action expert, so it is trained from random initialization jointly with the VLM backbone.

\textbf{Instruction Generalization Bottleneck.} Current continuous-action VLA policies perform well on seen tasks, but struggle with unseen instructions and compositional variations~\cite{black2024pi0,bjorck2025gr00tn1, kim2024openvla,kim2025oft}. Prior work traces this failure to a structural imbalance in VLA data: a trajectory of $T$ vision-action pairs shares a single language instruction, making visual-action diversity at least $T$ times richer than language~\cite{xu2025bayesvla,fang2026vision}. A randomly initialized action expert trained on such data learns to predict actions from visual cues alone, forming shortcuts that bypass language~\cite{xu2025bayesvla,lian2026langforce}(See Appendix~\ref{sec:inform-theory} for analysis). These shortcuts further produce noisy gradients that corrupt the VLM backbone, harming its language representations~\cite{driess2025knowledge, bjorck2025gr00tn15}. Just as VLMs gain generalizable capability by pretraining on balanced vision-language data~\cite{pertsch2025fast,black2025pi_ohfive,qwen3vl2025,beyer2024paligemma}, we argue that the action expert should likewise be pretrained on balanced action data to build coherent visuomotor priors free of shortcut incentives.

\subsection{Action Pretraining under Bayesian Formulation}
\label{sec:bayes}

To isolate the effect of language from action generation, we decompose the VLA policy via Bayesian factorization into a {Vision-Action~(VA) prior} and a {Language likelihood}:
\begin{equation}
  \pi(\mathbf{a} \mid \mathbf{v}, \ell)
    \;\propto\;
    {\pi^p(\mathbf{a} \mid \mathbf{v})}
    \cdot
    {\mathcal{L}(\ell \mid \mathbf{v}, \mathbf{a})}.
  \label{eq:bayes}
\end{equation}
\textbf{VA Prior.} The VA prior $\pi^p(\mathbf{a} \mid \mathbf{v})$ models the multimodal distribution of robot actions from visual observation, without any language input. It can be trained exclusively on vision-action pairs from the original VLA dataset. Every visual frame is paired with a unique action annotation, so the data is inherently balanced. Then, the action expert can develop an action manifold that captures diverse manipulation behaviors, without the risk of shortcut learning. Crucially, this produces a {meaningful pretraining} of the action expert before any language signal is introduced, directly addressing the bottleneck identified above.

\textbf{Language Likelihood.} With a well-initialized action expert from $\pi^p(\mathbf{a}\mid\mathbf{v})$, the likelihood $\mathcal{L}(\ell \mid \mathbf{v}, \mathbf{a})$ aligns the pretrained action distribution with the specific task instruction. In this paper, we do not separately parameterize $\mathcal{L}(\ell \mid \mathbf{v}, \mathbf{a})$. Instead, it is implicitly learned by injecting language condition into the action expert and jointly optimizing the whole policy. This preserves the action expert's learned priors while enabling effective language conditioning.

\textbf{Training Recipe.} The perspective in Eq.~\eqref{eq:bayes} motivates a two-stage pretraining procedure~(Figure~\ref{fig:overview}), followed by task-specific post-training:
\textbf{Pretraining Stage 1: VA Prior Pretraining.} The action expert is trained on one half of the pretraining data without language input, learning $\pi^p(\mathbf{a}\mid\mathbf{v})$. All VLM parameters are fixed. \textbf{Pretraining Stage 2: VLA Alignment.} The pretrained action expert is further conditioned on language input, then the whole model is trained on the full pretraining data, jointly optimizing $\pi^p(\mathbf{a}\mid\mathbf{v})$ and $\mathcal{L}(\ell \mid \mathbf{v}, \mathbf{a})$. \textbf{Post-Training: Task-specific Finetuning.} Similar as Pretraining Stage 2, the full policy is finetuned on task-specific data to adapt to target-domain distributions.

\subsection{Action Expert Design}
\label{sec:arch}

\begin{wrapfigure}{r}{8cm}
\centering
\vspace{-0.95cm}
\includegraphics[width=0.6\textwidth]{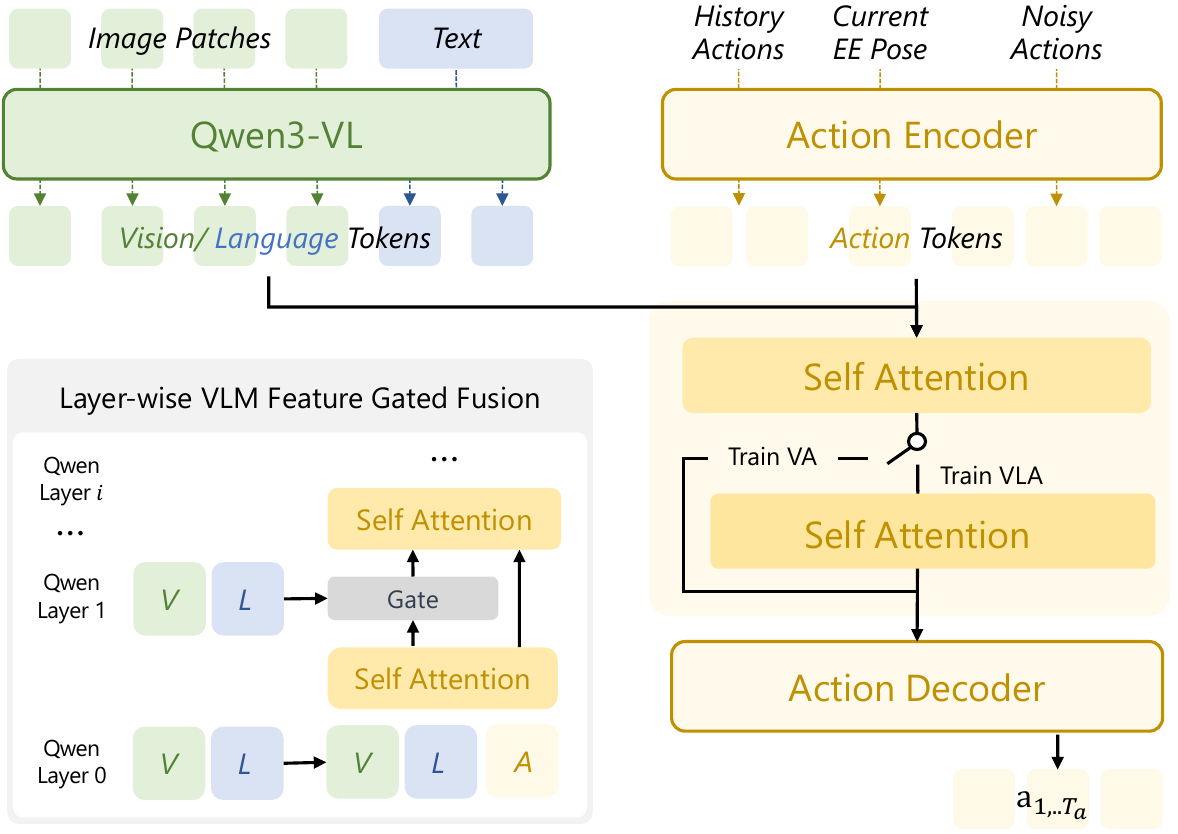}
\vspace{-0.35cm}
\caption{Action Expert Design. VLM features are injected into action expert via gated fusion. The action expert processes multimodal tokens by self-attention.}
\vspace{-0.55cm}
\label{fig:method}
\end{wrapfigure}

Figure~\ref{fig:method} shows the action expert design. 

\textbf{Multimodal Self-Attention.} The action expert is a Transformer-based diffusion model. At each denoising step, visual tokens $\mathbf{v}$, language tokens $\ell$, and action tokens $\mathbf{a}$ are concatenated into a single sequence and processed via block-wise causal self-attention. The action token sequence contains three parts: $\mathbf{a} = \bigl[\mathbf{a}^{\text{hist}},\;\mathbf{s}^{\text{prop}},\;
  \mathbf{a}^{\text{noisy}}\bigr]$,
where $\mathbf{a}^{\text{hist}}$ encodes executed action history, $\mathbf{s}^{\text{prop}}$ encodes current proprioceptive state, and $\mathbf{a}^{\text{noisy}}$ are noisy action tokens to be denoised. The denoising timestep is injected into each attention layer via Feature-wise Linear Modulation~(FiLM)~\cite{perez2018film}.

\textbf{Layer-wise VLM Feature Gated Fusion.} To leverage the semantic representations of large-scale VLMs, we use Qwen3-VL~\cite{qwen3vl2025} as our VLM backbone and inject its intermediate features into every self-attention layer of the action expert. For an action expert with $N$ attention layers, we uniformly sample $N$ intermediate features from Qwen3-VL at equal depth intervals. The vision and language input tokens $\mathbf{h}=[\mathbf{h}_v,\mathbf{h}_\ell]$ of the $(i{+}1)$-th attention layer of the action expert is:
\begin{equation}
  \mathbf{h}^{(i+1)}_{\text{in}}
    = \mathbf{h}^{(i)}_{\text{out}}
    + \sigma(\hat{w}_i)\cdot \phi_i^{\text{Qwen3-VL}}(\mathbf{v}, \ell), \\
  \label{eq:vlm_inject}
\end{equation}
$\hat{w}_i$ is a learnable scalar and $\sigma(\cdot)$ is the sigmoid function, acting as a gate that modulates the influence of each VLM layer on the action expert. For the first attention layer of the action expert, the input is directly the VLM's input embedding: $\mathbf{h}_\text{in}^{0}=\phi_0^{\text{Qwen3-VL}}(\mathbf{v}, \ell)$. This layer-wise gated fusion enables the action expert to assimilate both shallow spatial features and deep semantic features from Qwen3-VL, while maintaining its own vision-language pathway through the self-attention layers.

\textbf{Two Stage Pretraining.} The two training stages are realized within a single network via two complementary mechanisms. \textbf{Stage 1 VA Prior}: In this stage, we only activate $N/2$ attention layers of action expert for training. The language tokens from Qwen3-VL are completely masked from all self-attention computations in the action expert, reducing the network to a pure vision-action function: $[\mathbf{h}_v, \mathbf{h}_a]=\operatorname{SelfAttn}([\mathbf{v},\mathbf{a}])$.
\textbf{Stage 2 VLA Alignment}: After Stage 1 pretraining, we expand to the action expert from $N/2$ to $N$ attention layers by inserting an interleaved attention layer after each of the original $N/2$ layers. The language mask is removed, and all tokens participate in block-wise causal attention: $[\mathbf{h}_v, \mathbf{h}_\ell,\mathbf{h}_a]=\operatorname{SelfAttn}([\mathbf{v},\ell,\mathbf{a}])$.
The original $N/2$ layers are initialized from the Stage 1 checkpoint, leveraging the learned VA prior. The newly inserted $N/2$ layers then introduce language condition. Unlike BayesVLA~\cite{xu2025bayesvla} that freezes the prior in Stage 2, \ours jointly trains all $N$ layers on the full pretraining dataset for VLA alignment.

%% file: sec/experiment.tex

\section{Experiments}
\label{sec:result}
We conduct comprehensive experiments to validate: 1) the effectiveness of action expert pretraining; 2) to evaluate the instruction generalization of our policy; 3) to validate the architecture designs. 

\subsection{Simulation Experiments}

\subsubsection{Benchmarks}
We evaluate \ours mainly on two simulation benchmarks of language generalization difficulty:

\textbf{LIBERO-PRO}~\cite{zhou2025liberopro} introduces two perturbation types on top of LIBERO~\cite{liu2023libero}: (1)~{Pos}, which swaps object positions while fixing the instruction, testing whether the policy uses language rather than positional priors to locate targets; and (2)~{Task}, which replaces the manipulated object in the instruction with a different object in the same scene, forming an unseen task for OOD language generalization. Notably, simply replicating training trajectories fails under both perturbations, preventing dataset-level shortcuts.

\textbf{Rigid Object Pick-Place.} Following~\cite{xu2025bayesvla}, we evaluate on a language-conditioned pick-and-place benchmark with four suites: {Seen Object}~({SO}), {Unseen Object}~({UO}), {Unseen Container}~({UC}), and {Unseen Object \& Unseen Environment}~({UOUE}). Object layouts, camera viewpoints, and instructions are randomized across all suites, requiring genuine language following even in the SO setting.
\subsubsection{Baselines}
\label{subsec:baselines}

Our baselines include representative end-to-end VLA policies pretrained on large-scale datasets and finetuned on benchmark-specific data. These include \textbf{OpenVLA}~\cite{kim2024openvla,kim2025oft}, \textbf{$\pi_0$}~\cite{black2024pi0}, and \textbf{$\pi_{0.5}$}~\cite{black2025pi_ohfive}, where {OpenVLA} and {$\pi_0$} jointly train VLM and action expert, and {$\pi_{0.5}$} apply Knowledge Insulation~(KI)~\cite{driess2025knowledge} to stop the gradient from action expert to VLM during training. We also compare to \textbf{LangForce}~\cite{lian2026langforce}, a concurrent 
work that enforces language following by maximizing the mutual information between actions and instructions via a dual-branch architecture, and \textbf{CaP-X}, which generates structured task programs via a language 
model and executes them through learned robot primitives, providing an agent-level baseline. See Appendix~\ref{sec:baseline} for more details. 
For brevity, we use \textbf{\ours} to refer to both the action pretraining method and our action expert design. 

\subsubsection{Main Results}
\label{subsec:main_results}


\textbf{LIBERO-PRO.} Table~\ref{tab:libero_pro} reports results on LIBERO-PRO. OpenVLA and $\pi_0$ achieve 0\% success under both perturbations, confirming that direct joint training collapses to visual shortcuts and fails once language matters. $\pi_{0.5}$ recovers on Pos via KI, indicating effective preservation of in-distribution language following. However, it remains near zero on Task, showing KI alone cannot bridge the gap to OOD language generalization. LangForce improves over $\pi_{0.5}$ on Task, but degrades sharply on Pos across all suites, where the language is unchanged but object layouts shift. This may be because aggressively enforcing the action-language coupling makes the policy overly rely on language signals, suppressing the visual cues required for layout adaptation. This exposes a trade-off: $\pi_{0.5}$ preserves visuomotor robustness but fails on novel language, while LangForce gains language sensitivity but loses visual adaptability. \ours resolves this trade-off from the initialization side: Stage~1 VA prior training endows the action expert with visuomotor competence to handle layout variation, while Stage~2 VLA training adds language alignment without disrupting the pretrained distribution. As a result, \ours substantially outperforms both $\pi_{0.5}$ and LangForce, validating the effectiveness of action pretraining. \ours~(Ft VLM) further improves across most suites, showing that stop-gradient is not a necessary condition for language generalization: given a well-initialized action expert, jointly finetuning the VLM yields additional gains. One exception is Goal-Pos, where \ours scores below $\pi_{0.5}$: for most cases, \ours correctly follows the language, but may fail in obstacle avoidance, which is emphasized in this subsuite. Compared to CaP-X, \ours~(Ft VLM) achieves a higher average without an agent framework, showing that action pretraining enables a unified policy to match or exceed modular agent systems.

\begin{table}[t]
    \centering
    \begin{minipage}[t]{0.58\linewidth}
        \centering
        \scriptsize
        \renewcommand{\arraystretch}{0.87}
        \setlength{\tabcolsep}{3pt}
        \begin{tabular}{lcccccccccc}
            \toprule
            \multirow{2}{*}{Method}
                & \multicolumn{2}{c}{Spatial}
                & \multicolumn{2}{c}{Object}
                & \multicolumn{2}{c}{Goal}
                & \multicolumn{2}{c}{Long}
                & \multirow{2}{*}{Avg} \\
            \cmidrule(lr){2-3}\cmidrule(lr){4-5}\cmidrule(lr){6-7}\cmidrule(lr){8-9}
            & Pos & Task & Pos & Task & Pos & Task & Pos & Task & \\
            \midrule
            OpenVLA~\cite{kim2024openvla}          & 0 & 0 & 0 & 0 & 0 & 0 & 0 & 0 & 0 \\
            $\pi_0$~\cite{black2024pi0}            & 0 & 0 & 0 & 0 & 0 & 0 & 0 & 0 & 0 \\
            $\pi_{0.5}$~\cite{black2025pi_ohfive}  & 20 & 1 & 17 & 1 & \textbf{38} & 0 & \underline{8} & 1 & 11 \\
            LangForce~\cite{lian2026langforce}     & 11 & \underline{48} & 10 & 10 & 4 & 11 & 2 & \textbf{15} & 14 \\
            CaP-X~\cite{fu2026cap}                 & 12 & 14 & \underline{22} & \textbf{18} & \underline{26} & \underline{17} & -- & -- & -- \\
            \midrule
            \ours       & \underline{44} & \underline{48} & 7 & 10 & 23 & 11 & 6 & {3} & \underline{19} \\
            \ours~(Ft VLM)  & \textbf{62} & \textbf{62} & \textbf{24} & \underline{17} & 10 & \textbf{20} & \textbf{12} & \underline{12} & \textbf{27} \\
            \bottomrule
        \end{tabular}
        \vspace{0.2cm}
        \captionof{table}{Results on LIBERO-PRO (success rate~\%).}
        \label{tab:libero_pro}
    \end{minipage}
    \hspace{-3pt}%
    \begin{minipage}[t]{0.41\linewidth}
        \centering
        \scriptsize
        \renewcommand{\arraystretch}{1.2}
        \setlength{\tabcolsep}{2.6pt}
        \begin{tabular}{lccccccc}
            \toprule
            Method & KI & 2-Stage & Ft VLM & SO & UO & UC & UOUE \\
            \midrule
            $\pi_0$
                &  &  & \checkmark
                & 42 & 30 & 26 & 16 \\
            $\pi_{0.5}$
                & \checkmark & & \checkmark
                & 84 & 70 & 86 & 50 \\
            \midrule
            \multirow{4}{*}{\ours}
                &  &  & \checkmark
                & 88 & 56 & 66 & 34 \\
                & \checkmark &  &
                & 90 & 58 & 40 & 40 \\
                & \checkmark & \checkmark &
                & \underline{96} & \underline{74} & \underline{90} & \textbf{62} \\
                &  & \checkmark & \checkmark
                & \textbf{98} & \textbf{84} & \textbf{92} & \underline{58} \\
            \bottomrule
        \end{tabular}
        \vspace{0.2cm}
        \captionof{table}{Results on Pick-Place (rate~\%).}
        \label{tab:pickplace_results}
    \end{minipage}
    \vspace{-0.7cm}
\end{table}

\textbf{Rigid Object Pick-Place.} We report results in Table~\ref{tab:pickplace_results} across three design dimensions to trace the source of generalization gains: \textbf{KI}~(Knowledge Insulation~\cite{driess2025knowledge}), \textbf{2-Stage}~(action expert pretraining), and \textbf{Ft VLM}~(joint training of VLM and action expert). $\pi_0$~(no KI, Ft VLM) yields the weakest generalization. $\pi_{0.5}$ improves substantially by combining KI with large-scale VL data co-training. Among our four \ours variants trained solely on VLA data, adding KI without 2-Stage does not give obvious gains over the joint training variant~(Ft VLM), showing that stopping gradients alone does not resolve generalization failure. Combining KI with 2-Stage surpasses $\pi_{0.5}$ without any VL reasoning data, directly attributing the gain to action expert pretraining. Finally, replacing KI with joint VLM finetuning while retaining 2-Stage achieves the best overall result, confirming that KI is not a necessary condition. Instead, a well-initialized action prior allows joint VLM training to further improve rather than degrade language generalization.

\subsubsection{More Ablation Studies}
\label{subsec:ablation}

\begin{figure}[t]
  \centering
  \includegraphics[width=\textwidth]{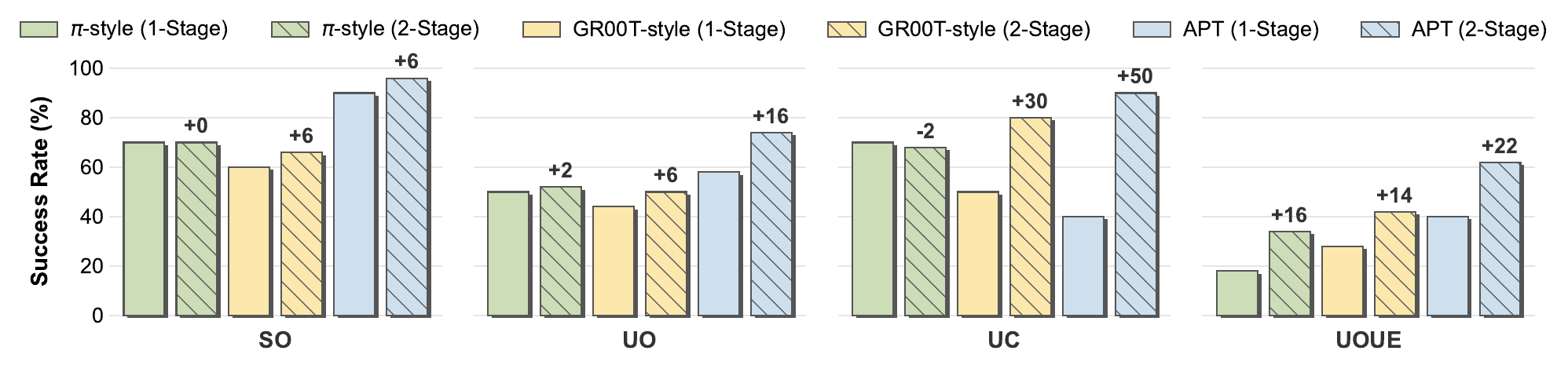}
  \vspace{-0.6cm}
  \caption{{Action expert pretraining applies to diverse architectures}.}
  \label{fig:ablation_two_stage}
\vspace{-0.6cm}
\end{figure}

\textbf{Action Pretraining on More Architectures.} To verify that action pretraining generalizes beyond our architecture design, we apply it to two representative architectures to combine VLM and the action expert: 
(1)~\textbf{$\pi$-style}~\cite{black2024pi0,black2025pi_ohfive}, where VLM and action expert tokens interact via block-wise causal self-attention at every attention layer and (2)~\textbf{GR00T-style}~\cite{bjorck2025gr00tn1,bjorck2025gr00tn15}, where only the final VLM layer feature conditions the action expert via cross-attention. Figure~\ref{fig:ablation_two_stage} shows that 2-Stage training improves generalization across almost all settings. The gain is most pronounced for \ours and GR00T-style compared to $\pi$-style. We attribute this to the VLM fusion design: \ours inserts VLM features by gated fusion, and GR00T-style uses only the final-layer features, both of which better preserve the VA prior learned in Stage~1. In contrast, $\pi$-style uses original VLM features at every self-attention layer, which less effectively preserves the pretrained action representations. 

\begin{wrapfigure}{r}{7cm}
\centering
\vspace{-0.6cm}
\includegraphics[width=0.5\textwidth]{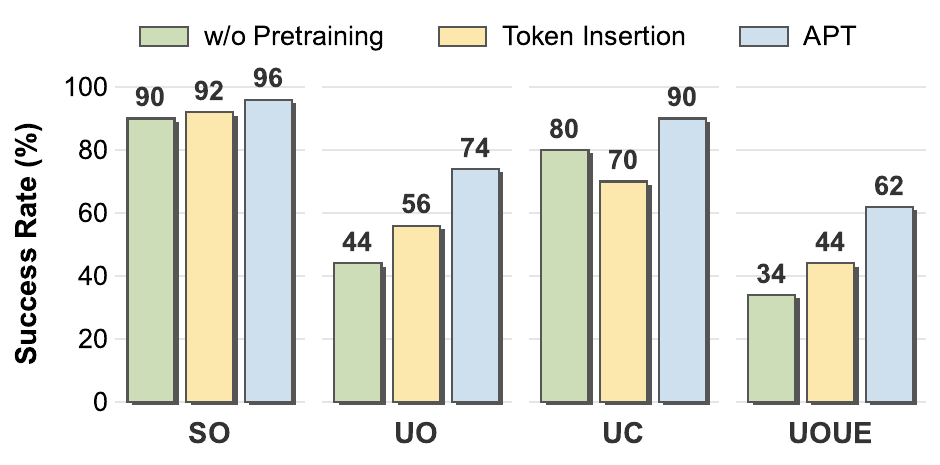}
\vspace{-0.6cm}
\caption{Ablation on large-scale pretraining and language injection mechanism.}
\vspace{-0.4cm}
\label{fig:ablation_pretrain}
\end{wrapfigure}

\textbf{Effectiveness of Large-Scale Pretraining.} To disentangle large-scale pretraining, we evaluate a variant applying two-stage training only on task-specific datasets, \textit{i.e.}, without any pretraining on large-scale datasets~(\textbf{w/o Pretraining}). As shown in Figure~\ref{fig:ablation_pretrain}, this variant still recovers meaningful generalization, confirming that action pretraining provides an inductive bias even in the low-data regime. However, it significantly underperforms \ours with pretraining on UO and UOUE, where generalizing to unseen categories and environments requires diverse action priors that can only be acquired from large-scale heterogeneous pretraining data.

\textbf{Language Injection Mechanism.} We compare our language injection mechanism via gated fusion against \textbf{Token Insertion}, which trains the VA prior with $N$ self-attention layers in Stage~1, and then plugs language tokens directly into the attention layers of the pretrained VA prior in Stage~2. Figure~\ref{fig:ablation_pretrain} shows that, our mechanism outperforms across all dimensions, with the largest gaps on UO and UOUE. \textbf{Token Insertion} abruptly modifies the pretrained VA distribution, risking partial forgetting of VA prior. Instead, through gated fusion that modulate language conditioning, the VA prior can be better preserved while enabling language-guided refinement of the action distribution.

\subsection{Real-World Experiments}
\label{sec:realworld}
 
\subsubsection{Experiment Setup}
\label{subsec:rw_setup}
In this section, we evaluate \ours in real-world setups. For each task, we collect 30 demonstrations to finetune each policy. We compare against $\pi_{0.5}$~\cite{black2025pi_ohfive} across all tasks. Our real-world evaluation covers two aspects: (1)~{Single Task Generalization}, which tests OOD language and object generalization within individual tasks; and (2)~{Compositional Task Generalization}, which evaluates sequential multi-task instruction following via task coaching (per-task instructions issued in sequence) and task chaining (a single concatenated instruction). See Appendix~\ref{sec:real-detail} for more details.

\subsubsection{Single Task Generalization}
\label{subsec:rw_task_gen}

\begin{figure}[t]
    \centering
    \begin{minipage}[t]{0.23\linewidth}
        \centering
        \includegraphics[width=\linewidth]{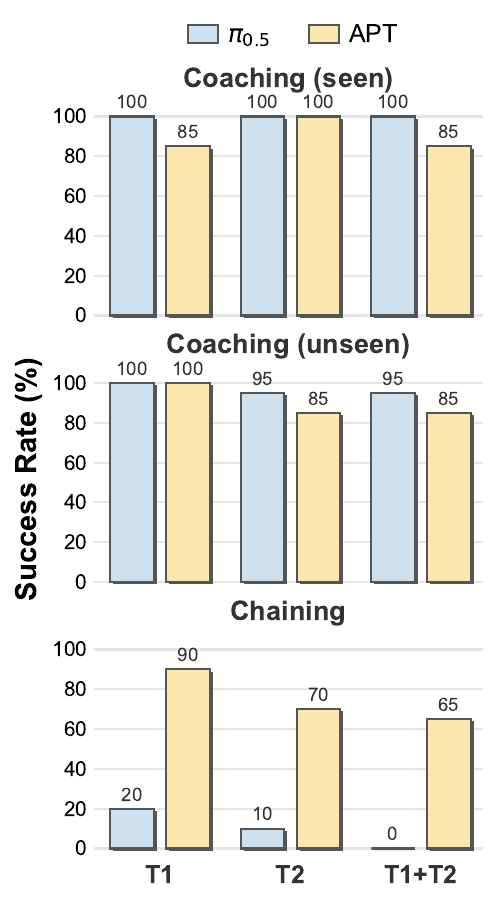}
        \vspace{-0.55cm}
        \captionof{figure}{Results on compositional task.}
        \label{fig:compositional}
    \end{minipage}%
    \hspace{4pt}%
    \begin{minipage}[t]{0.74\linewidth}
    \vspace{-5.8cm}   
        \centering
        \scriptsize
        \setlength{\tabcolsep}{3pt}
        \renewcommand{\arraystretch}{0.85}
        \setlength{\tabcolsep}{6pt}
        \begin{tabular}{l cccc cccc}
            \toprule
            \multirow{2}{*}{Method}
                & \multicolumn{4}{c}{Pick-Place}
                & \multicolumn{4}{c}{Clutter Pick-Place} \\
            \cmidrule(lr){2-5}\cmidrule(lr){6-9}
                & SO & UO & UOUC & UOUCUE
                & SO & UC & UO & UOUE \\
            \midrule
            $\pi_{0.5}$~\cite{black2025pi_ohfive}
                & 27/30 & 11/20 & 9/20 & 16/40
                & 18/30 & 18/30 & 4/10 & 3/10 \\
            \ours
                & \textbf{29/30} & \textbf{17/20} & \textbf{16/20} & \textbf{28/40}
                & \textbf{25/30} & \textbf{22/30} & \textbf{7/10} & \textbf{6/10} \\
            \bottomrule
        \end{tabular}
        \captionof{table}{Real-world task generalization results~(successes/trials).}
        \label{tab:real-task-gen}
        \vspace{3pt}
        
        \includegraphics[width=\linewidth]{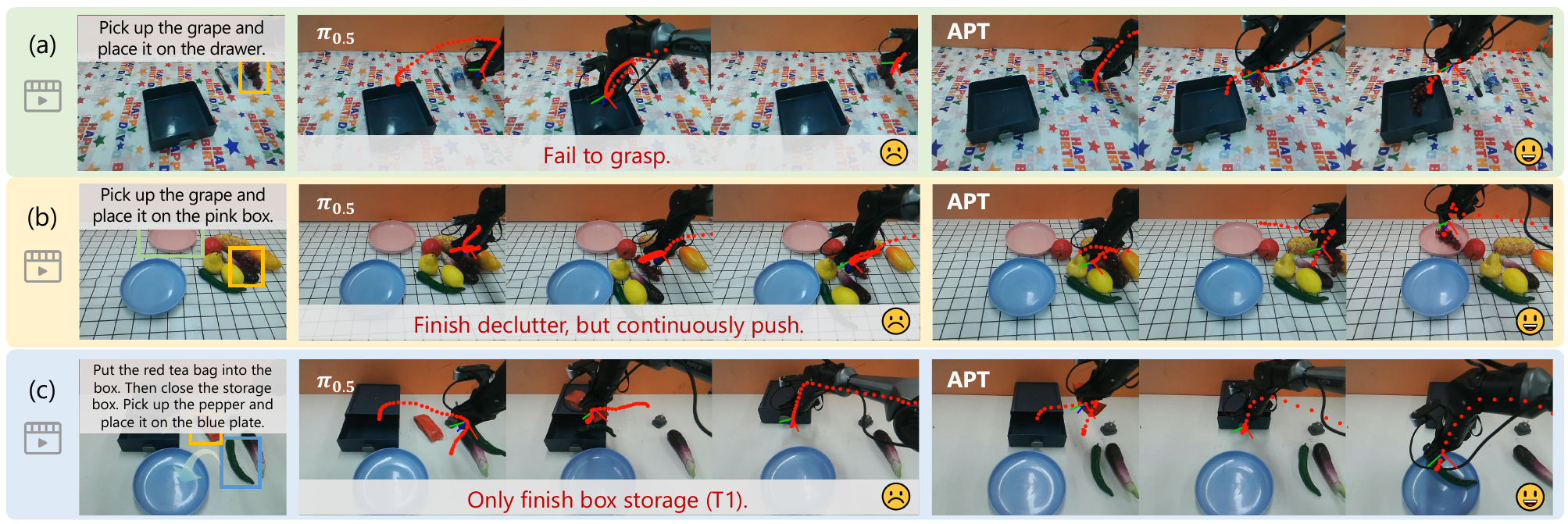}
        \vspace{-0.4cm}
        \captionsetup{margin={0.4cm, 0cm}}
        \captionof{figure}{Real-world cases. (a)~pick-place task, (b)~clutter pick-place task, (c)~compositional task chaining.}
        \label{fig:rw_comparison}
    \end{minipage}
    \vspace{-0.6cm}
\end{figure}

\textbf{Pick-Place Task}: The robot picks a specified object among several objects and places it on a specified container. We evaluate 110 trials across four settings of increasing OOD difficulty: seen objects~(SO), unseen objects~(UO), unseen objects with unseen containers~(UOUC), and further with unseen environments ~(UOUCUE). Each unseen setting demonstrates OOD language generalization. Table~\ref{tab:real-task-gen} shows that both methods achieve strong SO performance, confirming reliable seen language following. For OOD generalization, $\pi_{0.5}$ degrades significantly as task complexity increases. It sometimes fails to pick up the unseen target~(Figure~\ref{fig:rw_comparison}~(a)), or grasps a non-target object. In contrast, \ours maintains robust performance across all OOD levels. This is because the action expert is well initialized for instruction alignment, thus better inherits language generalization from VLM.

\textbf{Clutter Pick-Place Task}: The target object is tightly surrounded by distractors. The robot must first push to clear space before executing pick-and-place, requiring long-horizon planning across push, pick, and place sub-tasks~\cite{xu2021learning,chen2025clutterdexgrasp} and semantic grounding to identify the correct target. We evaluate over 80 trials across four settings (SO, UC, UO, UOUE). Results in Table~\ref{tab:real-task-gen} show that \ours outperforms $\pi_{0.5}$ across all settings, especially in OOD language generalization. $\pi_{0.5}$ sometimes struggles to transfer from push to grasp, or mistakenly pushes off the target. For unseen object settings, $\pi_{0.5}$ frequently grounds the wrong object when objects share similar colors~(\textit{e.g.}, misidentifying ``eggplant'' as ``grape''), and hesitates to grasp the target~(Figure~\ref{fig:rw_comparison}~(b)). \ours makes few such errors: validates that our action pretraining provides performance gain even in heavily cluttered long-horizon conditions.

\subsubsection{Compositional Task Generalization}
\label{subsec:rw_compositional}
We evaluate compositional instruction following across two tasks: {Table Storage~(T1)}, where the robot picks specified items from the table, places them in a storage box, and closes the box; and {Pick-Place~(T2)} from Section~\ref{subsec:rw_task_gen}.
Both $\pi_{0.5}$ and \ours achieve above 80\% success on each task in the seen setting, confirming sufficient individual task competence for the compositional evaluation. Notably, for the following evaluation, all the objects are seen during training. 
 
\textbf{Task Coaching}: Instructions for each task are issued sequentially as separate prompts~(seen or unseen language), with half in T1$\to$T2 order, half in T2$\to$T1 order. As shown in Figure~\ref{fig:compositional}, both methods perform comparably under seen language coaching, confirming that both retain strong language following when instructions are issued one at a time. For unseen language coaching, we perturb instructions via verb substitution, noun substitution, and cross-task object swapping. Both policies demonstrate robustness with nearly zero degradation. This indicates that single-task instruction understanding for seen objects is not the bottleneck for either method.
 
\textbf{Task Chaining}: A single concatenated prompt specifies both tasks jointly~(\textit{e.g.}, ``Put the red tea bag into the box. Then close the storage box. Pick up the pepper and place it on the blue plate.''). The results diverge dramatically: $\pi_{0.5}$ nearly collapses, while \ours maintains strong performance. This contrast reveals a key limitation of $\pi_{0.5}$: although it can follow individual instructions reliably, it fails to parse and execute multi-task instructions in a single prompt. It exhibits a typical failure mode of over-executing the first task: it attempts to place all target objects into the storage box, and fails to transition to the second task~(Figure~\ref{fig:rw_comparison}~(c)). Instead, \ours successfully executes chained tasks without explicit task segmentation, indicating better preservation of VLM's language understanding. Our failures arise from two sources: the policy occasionally advances to another task before the current task is fully completed, or it confuses the placement target with visually similar distractors.

%% file: sec/conclusion.tex
\section{Conclusion}
\label{sec:conclusion}

We propose action expert pretraining on balanced vision-action data to improve OOD language generalization in continuous-action VLA policies. Motivated by a Bayesian perspective, we first initialize a VA prior and then jointly finetune the policy with language conditioning. Comprehensive experiments validate that APT consistently improves instruction generalization, and applies to diverse architectures.

\textbf{Limitations.} Our current design does not explicitly model long-horizon memory, limiting generalization on tasks that require tracking multi-step progress. Besides, our evaluation focuses on tabletop manipulation. Extending to locomotion or mobile manipulation remains to be explored.

%% file: sec/acknowledgment.tex

\acknowledgments{This work is supported by the National Nature Science Foundation of China under Grant No. 62522317 and Zhejiang Provincial Natural Science Foundation of China under Grant No. LD25F030001.}


%% file: sec/appendix.tex
\appendix

\section{Implementation Details}
\label{sec:impl}

\begin{figure}[t]
  \centering
  \includegraphics[width=0.7\textwidth]{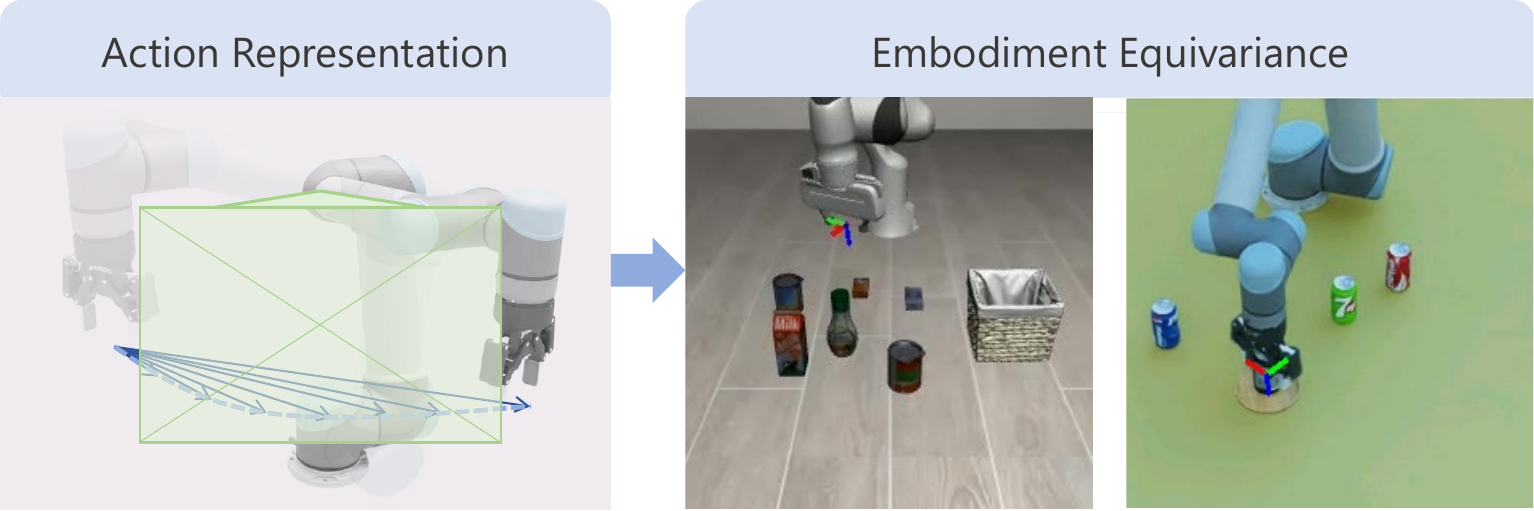}
  \caption{Action representation. We use relative end-effector poses projected in the camera frame, which is shared across embodiments.}
  \label{fig:action}
\vspace{-0.2cm}
\end{figure}

\textbf{Action Representation.} Actions are defined on the SE(3) manifold: 3D translation, 6D continuous rotation~\cite{zhou2019rot6d}, and normalized gripper width~($-1$ fully closed, $+1$ fully open), yielding a 10-dimensional action vector. Following~\cite{chen2025e2vla}, all end-effector poses and trajectories are expressed in the camera coordinate frame~(Figure~\ref{fig:action}). This representation disentangles the embodiment-relative information and provides embodiment equivariance across the heterogeneous platforms used for pretraining.

\textbf{Network Architecture.} The VLM backbone is Qwen3-VL-2B-Instruct~\cite{qwen3vl2025}. Visual tokens and language tokens are processed by separate MLP projectors before being injected into the action expert's embedding space. The action expert contains approximately 100M parameters, including $N=20$ attention layers. Both the action encoder and decoder are 2-layer MLPs with hidden dimension 768. We use an action chunk length of $T_a = 32$ and a history action length of $1$. Input images are resized to $256\times256$ for the wrist and third-view cameras. Proprioception is encoded as a 10-dimensional vector following the action representation above.

\begin{figure}[t]
  \centering
  \includegraphics[width=0.7\textwidth]{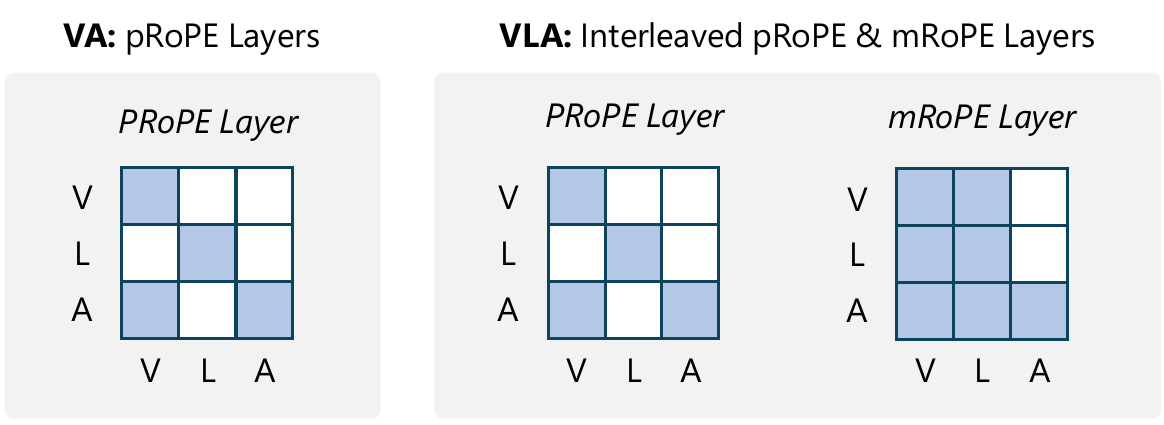}
  \caption{Attention masks of the two-stage training. Stage 1 uses PRoPE layers over vision and action tokens only. Stage 2 interleaves the inherited PRoPE layers with newly inserted mRoPE layers that jointly attend over vision, language, and action tokens.}
  \label{fig:attn_mask}
\vspace{-0.4cm}
\end{figure}

\textbf{Positional Encodings.} The two stages employ different positional encodings suited to their respective objectives. In Stage 1, visual and action tokens use Projective Positional Encoding~(PRoPE)~\cite{li2025cameras}, which embeds camera extrinsics directly into the positional embedding and enables multi-view spatial reasoning. In Stage 2, the inherited $N/2$ layers retain PRoPE, while the newly inserted $N/2$ layers use Multimodal Rotary Position Embedding~(mRoPE)~\cite{su2024roformer,qwen3vl2025} that natively handles interleaved vision and language tokens. This design lets the inherited layers maintain action understanding while the inserted layers introduce language condition for multimodal alignment. Figure~\ref{fig:attn_mask} illustrates the attention masks of the two layer types across both stages.

\textbf{Training Settings.} We use the AdamW optimizer, with action expert learning rate $10^{-4}$ and weight decay $10^{-2}$. When jointly finetuning the VLM, we use a VLM learning rate of $10^{-5}$ and weight decay of $10^{-10}$. Models are trained with a global batch size of 256 on NVIDIA A100 80G. Stage 1 and Stage 2 are each trained for 100k iterations. Note that in Stage 1, Qwen3-VL receives both vision and language tokens, but a block-wise causal mask prevents vision tokens from attending to language tokens, while the action expert also masks language, thus avoiding language leakage. We adopt DDPM~\cite{ho2020ddpm} with 100 diffusion steps for training, and DDIM~\cite{songdenoising} with 20 denoising steps for inference.

\textbf{Pretraining Datasets.} We pretrain on four large-scale datasets covering diverse embodiments, scenes, and skills.
\begin{itemize}[leftmargin=*, topsep=2pt, itemsep=1pt]
    \item \textit{DROID}~\cite{khazatsky2024droid} contains 78{,}544 real-world Franka trajectories~(about 350 hours) across 564 scenes and 84 task categories with multi-view RGB-D and end-effector pose labels.
    \item \textit{AgiBotWorld-Alpha}~\cite{bu2025agibotworld} provides over one million real-world dual-arm trajectories spanning 217 tasks across five deployment scenarios.
    \item \textit{InternData-A1}~\cite{tian2025internA1} is a high-fidelity synthetic dataset with about 630k trajectories across 4 embodiments, 18 skills, 70 tasks, and 227 scenes, spanning rigid, articulated, deformable, and fluid manipulation.
    \item \textit{InternVLA-M1}~\cite{chen2025internvlam1} contributes 244k generated pick-place trajectories across 200 tasks and over 3{,}000 object instances, providing dense language-object correspondences.
\end{itemize}
During pretraining, we sample the four datasets with weights $5$:$5$:$4$:$1$ for DROID, AgiBotWorld-Alpha, InternData-A1, and InternVLA-M1 respectively, balancing real-robot diversity with synthetic skill coverage.

\section{Visual Shortcut Analysis}
\label{sec:inform-theory}
Following~\cite{xu2025bayesvla}, we use information theory to validate that the standard VLA training objective admits a shortcut solution that largely ignores language, and that our two-stage training mitigates this issue. The analysis is consistent with our main claim: a randomly initialized action expert, trained jointly with the VLM, falls into the shortcut degrading OOD language generalization.

Formally, minimizing the VLA objective corresponds to minimizing the conditional negative log-likelihood of actions given vision and language, which lower-bounds at the conditional entropy:
\begin{equation}
      \min -\mathbb{E} [\log\pi(\mathbf{a}|\mathbf{v},\ell)] \approx \mathcal{L}_\mathrm{VLA} = H(\mathbf{a}|\mathbf{v},\ell)
\end{equation}
When training converges, the negative log-likelihood objective approximates this entropy lower bound.


{\bf Dataset Imbalance.} In most VLA datasets, language diversity is much smaller than vision-action diversity. A trajectory typically consists of many visual frames associated with a single language prompt. Although two trajectories with different prompts may share a small number of visually similar frames, such overlaps occur only rarely. As a result, for most visual observations, the corresponding language instruction is nearly deterministic. Formally, denoting $P(\ell|\mathbf{v})$ as the probability of language prompt conditioned on the visual frame, we state the assumption that $P(\ell|\mathbf{v})$ is effectively one-hot for the majority of frames. In other words, the language can often be inferred directly from the visual observation. Consequently, the conditional entropy is upper-bounded by a small constant, $H(\ell|\mathbf{v}) \leq \epsilon$. For example, if a drawer is visible in a frame, the associated instruction is highly likely to be opening or closing the drawer. Such imbalance is prevalent in existing embodied datasets and implies that language provides limited additional information beyond what is already encoded in the visual input.


\subsection{Shortcut Learning}
Under this assumption, we show that the standard VLA training admits a shortcut solution. Defining the conditional mutual information $I$ as
\begin{equation}
\begin{aligned}
    I(\mathbf{a};\ell|\mathbf{v}) 
    & = H(\mathbf{a}|\mathbf{v}) - H(\mathbf{a}|\mathbf{v},\ell) \\
    & = H(\ell|\mathbf{v}) - H(\ell|\mathbf{v},\mathbf{a})
\end{aligned}
\end{equation}
where $H(\cdot)\geq0$. Combining with the dataset assumption gives
\begin{equation}
    0 \leq I(\mathbf{a};\ell|\mathbf{v}) \leq H(\ell|\mathbf{v})  \leq \epsilon
\end{equation}
leading to
\begin{equation}
    0 \leq  H(\mathbf{a}|\mathbf{v}) - H(\mathbf{a}|\mathbf{v},\ell)   \leq \epsilon
    \label{avavl}
\end{equation}
which means that the inclusion of the language prompt contributes only marginally to action prediction.

Now consider an alternative vision-only model trained with the objective
\begin{equation}
    \min -\mathbb{E}[\log\pi(\mathbf{a}|\mathbf{v})]
    \label{vamodel}
\end{equation}
whose optimal lower bound $\mathcal{L}_\mathrm{VA}$ is the conditional entropy $H(\mathbf{a}|\mathbf{v})$. Then we have
\begin{equation}
    \mathcal{L}_\mathrm{VLA} \le \mathcal{L}_\mathrm{VA} \leq \mathcal{L}_\mathrm{VLA} +\epsilon
    \label{eq:loss}
\end{equation}
This bound shows that a vision-only policy achieves only marginally larger loss than the full VLA policy. Since the vision-only policy is simpler~(it does not depend on $\ell$), gradient descent on the VLA objective from a random action expert is biased toward this vision-only solution. We refer to this as the \textit{visual shortcut}, and it explains why standard VLA training is weak at OOD instruction following.


\subsection{Two-stage Conditioning}
Bayesian factorization induces a two-stage training procedure that avoids this shortcut. In Stage 1, we learn the prior $\pi^p(\mathbf{a}|\mathbf{v})$ following Eqn.~\ref{vamodel}. Because the prior does not condition on language by construction, it is exactly the desired vision-action policy rather than a shortcut. Denoting the learned parameters as $\theta_\mathrm{VA}$,
\begin{equation}
    \mathcal{L}_\mathrm{VA} \approx \min -\mathbb{E}[\log \pi^p_{\theta_\mathrm{VA}}(\mathbf{a}|\mathbf{v})] 
\end{equation}
Let $f_{\theta_\mathrm{VA}}(\mathbf{v})$ denote the intermediate visual embedding produced by the pretrained prior. In Stage 2, we train the language-conditioned policy on top of this embedding:
\begin{equation}
    \mathcal{L}_\mathrm{VLA} \leq \min -\mathbb{E}[\log\pi(\mathbf{a}|f_{\theta_\mathrm{VA}}(\mathbf{v}),\ell)] \leq \mathcal{L}_\mathrm{VA}
    \label{stage2}
\end{equation}
where $\theta_\mathrm{VA}$ is initialized from Stage 1 and the newly added language-conditioning parameters are randomly initialized. If language were entirely ignored, the right inequality in Eqn.~\ref{stage2} would be tight at $\mathcal{L}_\mathrm{VA}$. Since the loss can be further optimized, the gradient incentivizes the network to use language for action generation. Therefore, the resulting policy can effectively ground the language prompt.


\section{Baseline Details}
\label{sec:baseline}
\textbf{OpenVLA}~\cite{kim2024openvla} is built on the Prismatic 7B VLM backbone~\cite{karamcheti2024prismatic}, pretrained on diverse datasets including Open X-Embodiment~\cite{openxembodiment}, and finetuned using the OpenVLA-OFT recipe~\cite{kim2025oft} with continuous action representations.

\textbf{$\pi_0$}~\cite{black2024pi0} combines PaliGemma~\cite{beyer2024paligemma} with a diffusion-based action expert, pretrained on a subset of OXE and the $\pi$ dataset, and finetuned on DROID~\cite{khazatsky2024droid}.

\textbf{$\pi_{0.5}$}~\cite{black2025pi_ohfive} extends $\pi_0$ with a hybrid pretraining procedure that jointly trains on internet-scale reasoning data and large-scale robot demonstration data, representing the current state of the art in generalist VLA policies.

\textbf{LangForce}~\cite{lian2026langforce} introduces latent action queries into Qwen3-VL~\cite{qwen3vl2025} and enforces language following by estimating vision-only and language-conditioned action distributions via dual branches and maximizing the mutual information between actions and instructions.

\textbf{CaP-X}~\cite{fu2026cap} is a code-as-policies framework that generates structured task programs via a language model and executes them through learned robot primitives, providing a complementary language-grounding baseline.

The following are additional baselines involved in the appendix:

\textbf{$\pi_0$-FAST}~\cite{pertsch2025fast} is a discrete-token variant of $\pi_0$ that replaces the diffusion action expert with a frequency-space action tokenization, enabling faster autoregressive action prediction.

\textbf{UniVLA}~\cite{bu2025univla} learns a unified task-centric latent action space from heterogeneous human videos and robot data, supporting cross-embodiment pretraining with a discrete latent action codebook.

\textbf{WorldVLA}~\cite{cen2025worldvla} couples a VLA policy with a world model that predicts future frames as an auxiliary objective to enhance visual representation learning.

\textbf{RIPT-VLA}~\cite{tan2025ript} introduces a reinforcement-learning-based interactive post-training stage that finetunes pretrained VLA models using sparse binary success rewards via dynamic rollout sampling and leave-one-out advantage estimation.

\textbf{X-VLA}~\cite{zheng2025xvla} is a flow-matching-based VLA architecture that uses soft prompts with embodiment-specific learnable embeddings to handle heterogeneous cross-embodiment data, built on standard transformer encoders for scalability.

\section{Simulation Experiment Details}
\subsection{Benchmark Details}
\label{subsec:benchmarks}

\begin{figure}[t]
    \centering
    \includegraphics[width=\linewidth]{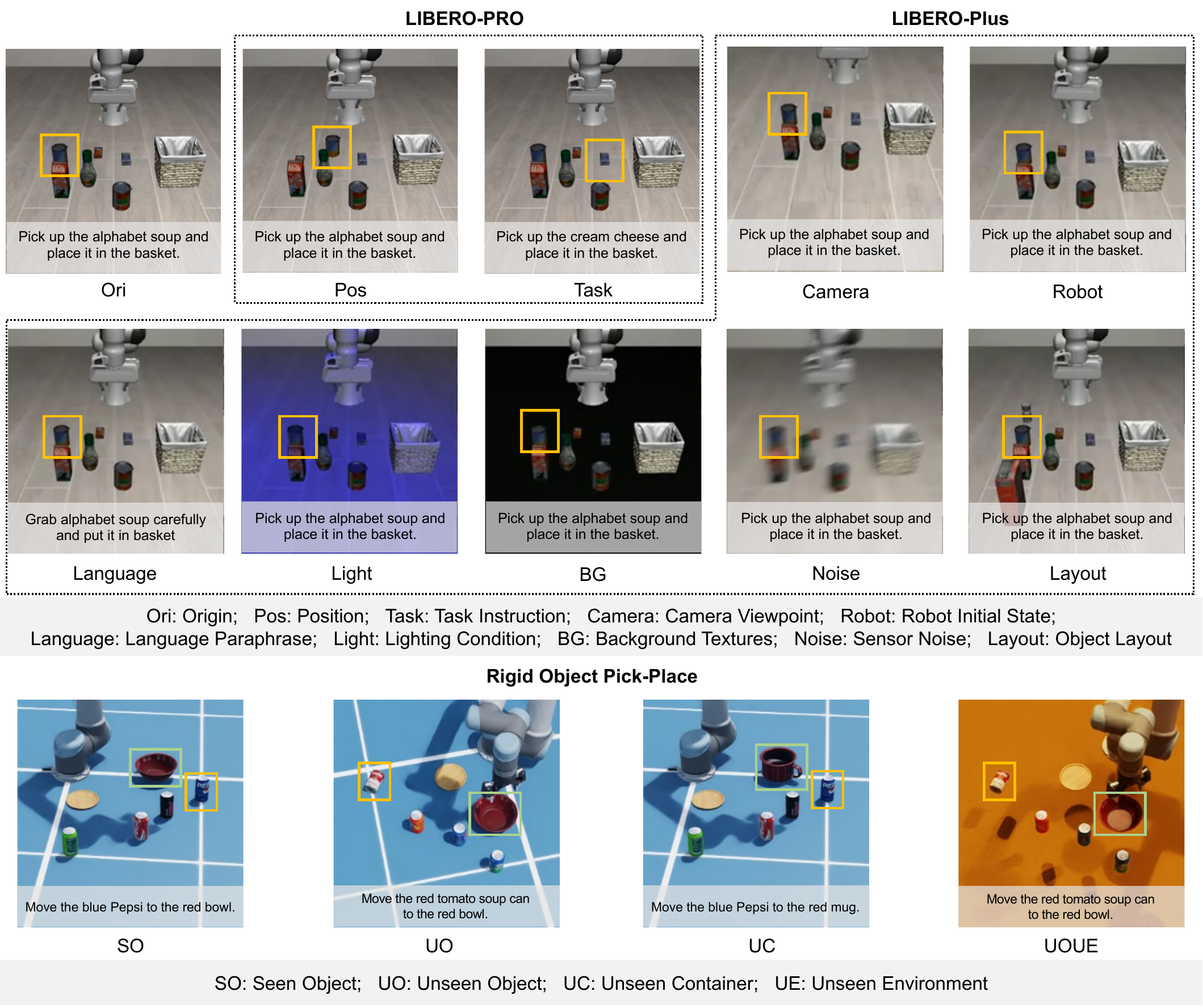}
    \caption{
        Overview of the simulation benchmarks.
    }
    \label{fig:sim_bench}
    \vspace{-0.3cm}
\end{figure}

We evaluate \ours on simulation benchmarks of increasing language-generalization difficulty. Figure~\ref{fig:sim_bench} summarizes the benchmark settings.

\textbf{LIBERO}~\cite{liu2023libero}. We use all four task suites of LIBERO. LIBERO-Spatial tests spatial relational understanding under fixed object sets; LIBERO-Object evaluates the ability to discriminate object instances under fixed scene layouts; LIBERO-Goal measures goal-directed instruction following; and LIBERO-Long contains 10 long-horizon sequential tasks. Each suite contains 10 tasks with 50 demonstrations each, yielding 500 trajectories per suite. Notably, LIBERO is an in-distribution evaluation rather than a generalization benchmark: training and evaluation share the same task set, object set, and language template. 

\textbf{LIBERO-Plus}~\cite{fei2025liberoplus}. LIBERO-Plus extends LIBERO with seven axes of controlled perturbation: object layout, camera viewpoint, robot initial state, language instruction paraphrase, lighting conditions, background textures, and sensor noise. Five of them~(camera viewpoint, robot initial state, lighting conditions, background textures, and sensor noise) test robustness to visual and configuration changes while keeping the task and language fixed. The language axis paraphrases the original instruction without changing the target task, and primarily measures robustness to instruction wording. The object layout axis randomizes the initial positions of the manipulated objects, which probes both robustness to spatial distractors and the policy's ability to follow language under varied object configurations. This axis is therefore partially aligned with the language-following dimension we emphasize. Overall, however, LIBERO-Plus does not evaluate OOD task generalization, since the target task semantics remain unchanged across all seven perturbations.


\textbf{LIBERO-PRO}~\cite{zhou2025liberopro}. To explicitly evaluate language generalization under controlled perturbations, we adopt LIBERO-PRO, which introduces two structured perturbation types on top of LIBERO: (1)~{Pos}, which swaps the initial positions of manipulated objects while holding the task instruction fixed, testing whether the policy genuinely uses language to locate the target rather than relying on positional priors; and (2)~{Task}, which replaces the manipulated object in the instruction with a different object in the same scene, forming an unseen task for OOD language generalization. Notably, simply replicating training trajectories fails under both perturbations, preventing dataset-level shortcuts.

{\bf Rigid Object Pick-Place.} Following~\cite{xu2025bayesvla}, we evaluate on a simulation benchmark built in IsaacSim~\cite{NVIDIA_Isaac_Sim} for diverse language-conditioned pick-and-place tasks using a UR5 robot arm. Each scene randomly samples 4 of 25 objects covering diverse colors, shapes, sizes, and categories, with a randomly selected pick target. The place target is randomly chosen between a yellow plate and a red bowl. There are 10k pick-place trajectories across 500 scenarios for multi-task training, each with randomized camera viewpoints and object layouts. There are four evaluation suites: Seen Object~(SO), Unseen Object~(UO), Unseen Container~(UC), and Unseen Object \& Unseen Environment~(UOUE). SO uses seen pick objects and containers but with randomly sampled instructions, viewpoints, and layouts. UO samples 4 of 10 unseen objects~(with unseen colors, shapes, sizes, and categories) while retaining seen containers, testing generalization across object categories. UC uses seen objects with unseen containers, testing placement generalization. UOUE applies unseen backgrounds and lighting on top of UO. Language instructions, camera viewpoints, and object layouts are randomly sampled across all evaluations. 

\subsection{Results on Original LIBERO}

Table~\ref{tab:libero_results} reports results on the original LIBERO benchmark. \ours achieves an average success rate of $96.1\%$, with $98.4\%$ on Spatial, $99.4\%$ on Object, $96.4\%$ on Goal, and $90.2\%$ on Long. Since LIBERO primarily measures in-distribution task completion rather than OOD generalization, these results are not intended to claim superiority; rather, they confirm that \ours does not sacrifice task-solving capability. \ours remains competitive with $\pi_{0.5}$, and exceeds $\pi_0$, UniVLA on training-distribution tasks while substantially improving language generalization on the more challenging benchmarks reported in the main paper~(Table~\ref{tab:libero_pro}).

\begin{table}[t]
    \centering
    \begin{tabular}{lccccc}
        \toprule
        Method & Spatial & Object & Goal & Long & Avg \\
        \midrule
        UniVLA~\cite{bu2025univla}              & 96.5 & 96.8 & 95.6 & \underline{92.0} & {95.2} \\
        WorldVLA~\cite{cen2025worldvla}         & 87.6 & 96.2 & 83.4 & 60.0 & 81.8 \\
        $\pi_0$~\cite{black2024pi0}             & {96.8} & \underline{98.8} & {95.8} & 85.2 & 94.2 \\
        $\pi_{0.5}$~\cite{black2025pi_ohfive}   & \textbf{98.8} & 98.2 & \textbf{98.0} & \textbf{92.4} & \textbf{96.9} \\
        \midrule
        \ours                                    & \underline{98.4} & \textbf{99.4} & \underline{96.4} & 90.2 & \underline{96.1} \\
        \bottomrule
    \end{tabular}
    \vspace{0.2cm}
    \caption{
        Results on LIBERO (success rate~$\%$).
        \textbf{Bold}: best. \underline{Underline}: second best.
    }
    \vspace{-0.2cm}
    \label{tab:libero_results}
\end{table}

\subsection{Results on LIBERO-Plus}

Table~\ref{tab:liberoplus} reports results on LIBERO-Plus across the seven perturbation axes. \ours achieves the highest average success rate, slightly above X-VLA, and clearly ahead of $\pi_0$ and UniVLA. \ours obtains the best result on object layout ($80.1\%$) and lighting ($93.6\%$), and the second-best result on language ($77.6\%$), background ($92.3\%$), and robot initial state ($63.1\%$). The combination results of layout~(best) and language~(second-best) axes indicate that \ours retains reliable instruction following under perturbations that typically expose visual-shortcut policies, which further supports the effectiveness of action expert pretraining.

\begin{table}[t]
    \centering
    \begin{tabular}{lcccccccc}
        \toprule
        Model & Camera & Robot & Lang. & Light & BG & Noise & Layout & Total \\
        \midrule
        $\pi_0$~\cite{black2024pi0}            & 13.8 &  6.0 & 58.8 & 85.0 & 81.4 & \textbf{79.0} & 68.9 & 53.6 \\
        $\pi_0$-FAST~\cite{pertsch2025fast}        & \textbf{65.1} & 21.6 & 61.0 & 73.2 & 73.2 & 74.4 & 68.8 & 61.6 \\
        OpenVLA~\cite{kim2025oft}                          & \underline{55.6} & 21.7 & \textbf{81.0} & \underline{92.7} & 91.0 & \underline{78.6} & 68.7 & 67.9 \\
        UniVLA~\cite{bu2025univla}              &  1.8 & 46.2 & 69.6 & 69.0 & 81.0 & 21.2 & 31.9 & 42.9 \\
        RIPT-VLA~\cite{tan2025ript}                                & 55.2 & 31.2 & \underline{77.6} & 88.4 & 91.6 & 73.5 & \underline{74.2} & 68.4 \\
        X-VLA~\cite{zheng2025xvla}                                   & 23.4 & \textbf{89.7} & 75.7 & 88.2 & \textbf{96.0} & 62.7 & 71.8 & \underline{71.4} \\
        \ours                                   & 31.8 & \underline{63.1} & \underline{77.6} & \textbf{93.6} & \underline{92.3} & 76.0 & \textbf{80.1} & \textbf{71.6} \\
        \bottomrule
    \end{tabular}
    \vspace{0.2cm}
    \caption{
        Results on LIBERO-Plus (success rate~$\%$) across seven robustness perturbation axes.
    }
    \vspace{-0.5cm}
    \label{tab:liberoplus}
\end{table}

\subsection{Case Studies on Rigid Object Pick-Place}
\label{subsec:case_studies}

\begin{figure}[t]
    \centering
    \includegraphics[width=\linewidth]{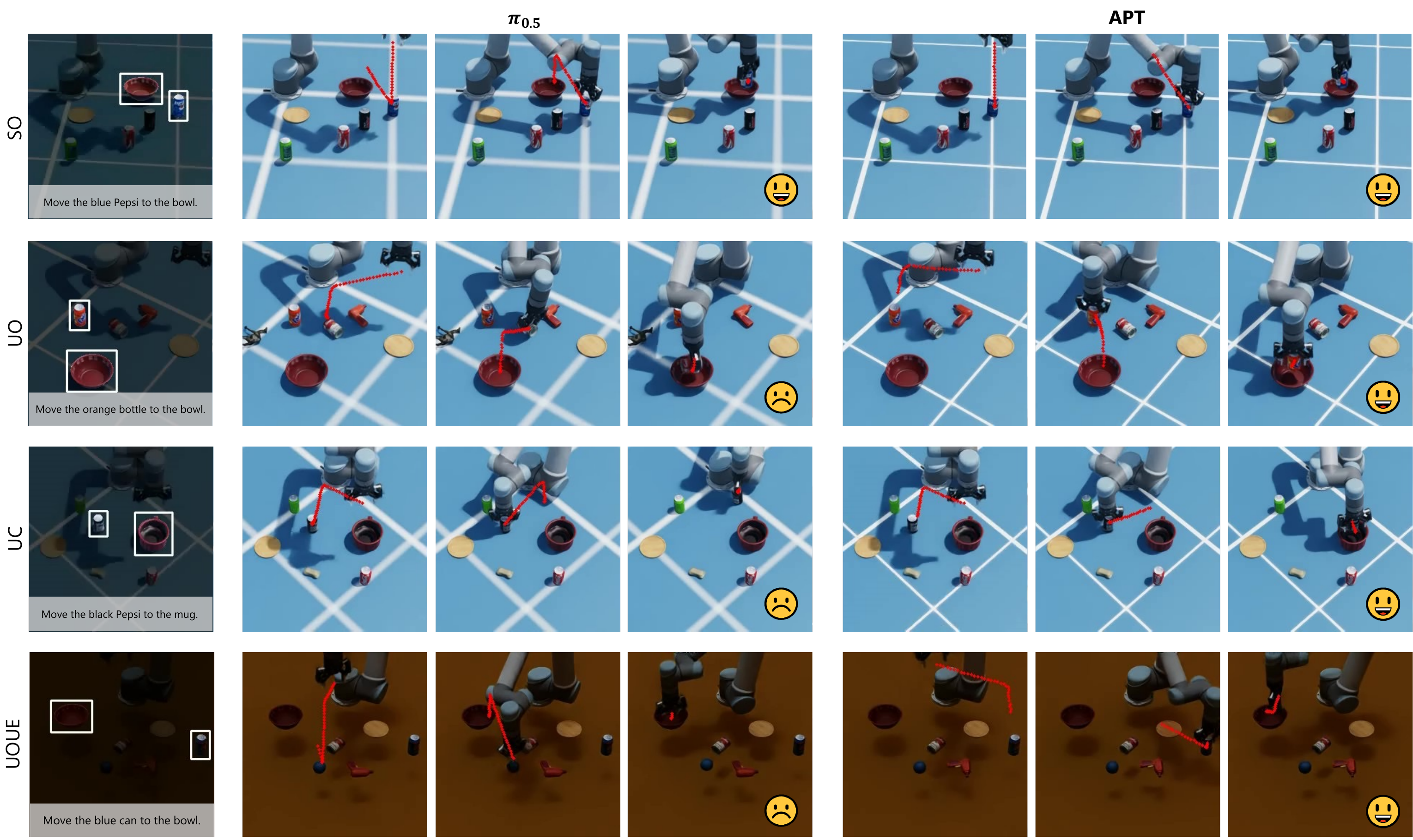}
    \caption{
        Qualitative case studies on rigid object pick-place across the four evaluation settings. Red dotted lines visualize end-effector trajectories. Annotated failure causes are highlighted.
    }
    \label{fig:sim_case}
    \vspace{-0.2cm}
\end{figure}

Figure~\ref{fig:sim_case} presents qualitative examples from the rigid object pick-place benchmark across all four evaluation splits. In the SO split, both $\pi_{0.5}$ and \ours follow the seen instruction reliably and grasp the correct object, indicating that in-distribution language following is largely solved for both methods. In the UO split, $\pi_{0.5}$ grasps a distractor whose color is similar to the language-specified target, while \ours correctly grounds the unseen object, showing that action expert pretraining preserves the VLM's semantic grounding for novel object categories. In the UC split, $\pi_{0.5}$ reaches the correct pick object but hesitates during placement and drops the object outside the unseen container, whereas \ours produces a precise placement trajectory aligned with the unseen container. In the most challenging UOUE split, $\pi_{0.5}$ frequently misgrasps distractors that share color or shape with the target, while \ours grasps the correct object more often. Together, these cases are consistent with the quantitative results in the main paper: action expert pretraining narrows the gap between the VLM's semantic generalization and the policy's grounded action, particularly under compound shifts in object identity and environment.

\section{Real-world Experiment Details}
\label{sec:real-detail}

\subsection{Real-world Setup}
\begin{figure}[t]
    \centering
    \includegraphics[width=0.6\linewidth]{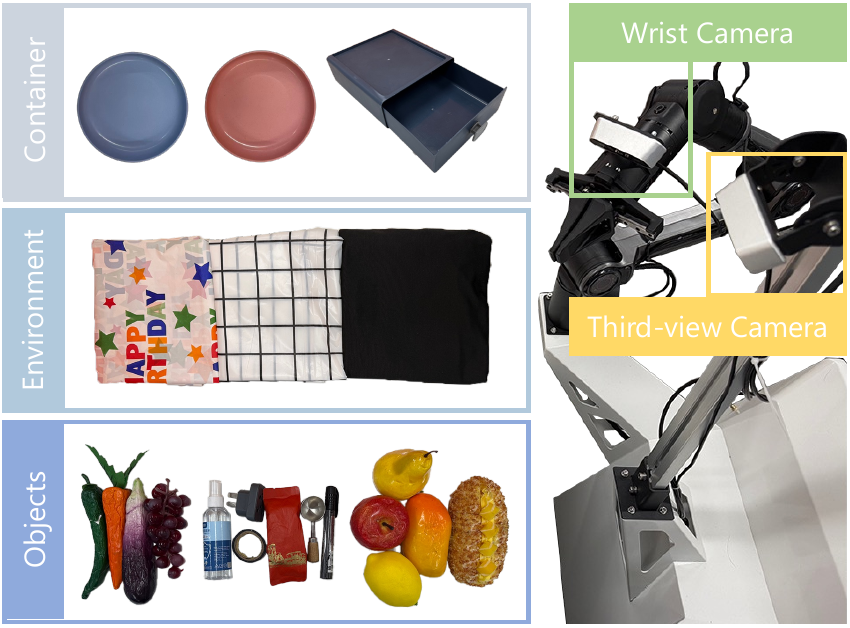}
    \caption{
        Real-world platform, containers, objects, and background variations used in our real-world evaluations. 
    }
    \label{fig:real_platform}
    \vspace{-0.3cm}
\end{figure}

We use the Agilex Cobot platform, equipped with a Piper robot arm and two ORBBEC DaBai cameras for third-view and wrist-view, both capturing RGB-D images at $640\times480$ resolution. Figure~\ref{fig:real_platform} shows the platform together with the tested objects, containers, and background variations.

For each task, we collect 30 tele-operated demonstrations covering a diversity of language instructions and object layouts. We compare \ours with $\pi_{0.5}$, which is the strongest baseline in our simulation experiments. In all real-world experiments, $\pi_{0.5}$ is finetuned using joint-space actions, consistent with its original pretraining setup, while \ours uses the camera-frame action representation from Section~\ref{sec:impl}.

\begin{figure}[t]
    \centering
    \includegraphics[width=\linewidth]{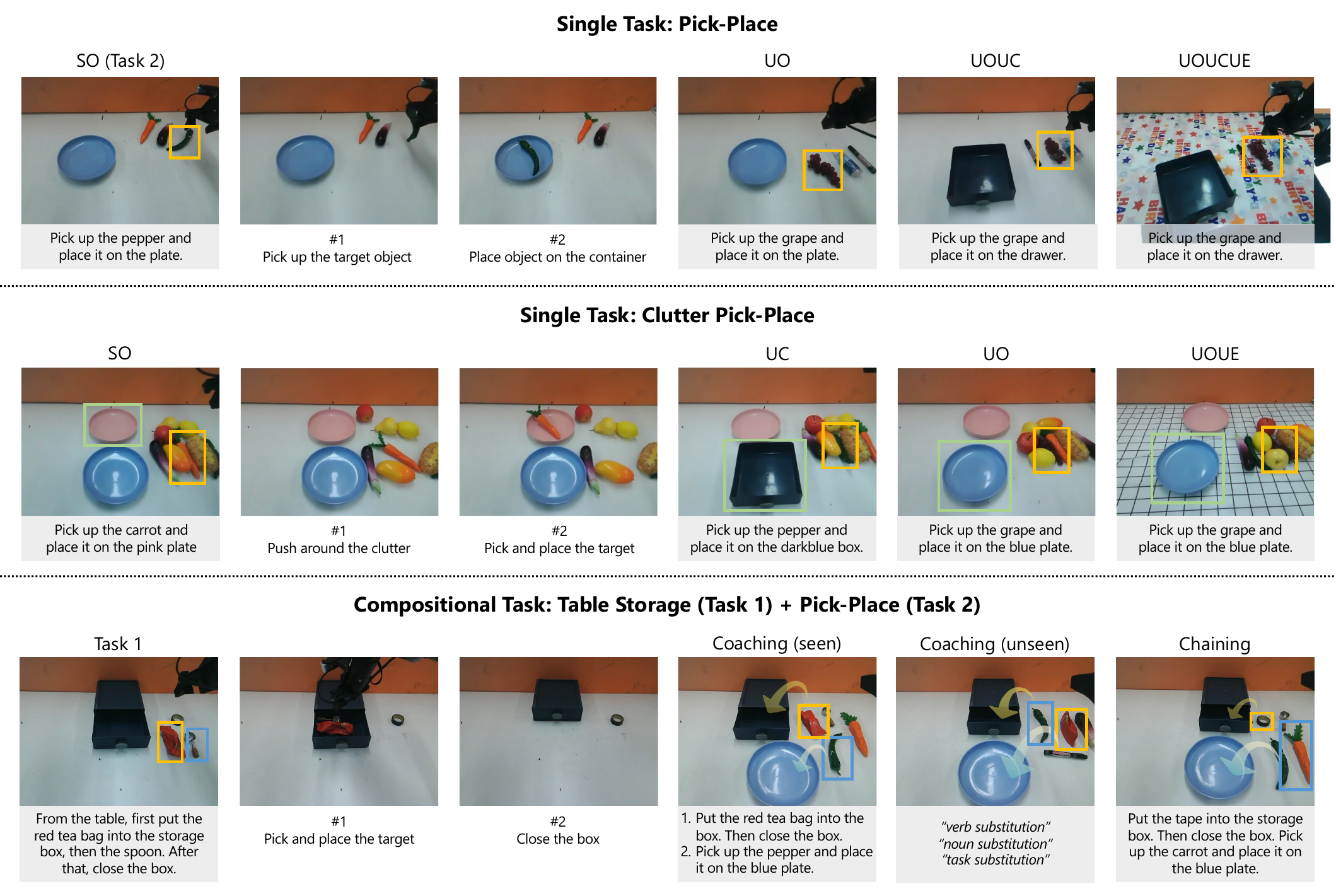}
    \caption{
        Real-world task and generalization setting overview. Top: single pick-place task with SO, UO, UOUC, and UOUCUE splits. Middle: single clutter pick-place task with SO, UC, UO, and UOUE splits. Bottom: compositional task combining table storage (T1) and pick-place (T2), evaluated under seen and unseen language coaching as well as task chaining.
    }
    \label{fig:real-bench}
    \vspace{-0.5cm}
\end{figure}

\subsection{Single Task Detailed Settings}
\textbf{Pick-place Task.} For this task, the policy is supposed to pick up the specified object into a container. The language instruction for this task is: ``Pick up the \{object\} and place it on the plate''. Representative successful rollout and example cases of different generalization dimensions are shown in Figure~\ref{fig:real-bench}. Each policy is evaluated over a total of 110 trials. In the first 30 trials, we evaluate the pick-place performance with the seen objects and container~(SO). For each target object, we conduct 10 trials, where we vary object positions, with other objects serving as distractors. This setup evaluates the language-following capability of policies. The remaining 80 trials evaluate the policy’s generalization performance across multiple dimensions. With two unseen objects as target items, we gradually increase the level of generalization: unseen object with seen container~(UO), unseen object with unseen container~(UOUC), and unseen object with unseen container under two distinct unseen backgrounds~(UOUCUE). Each configuration is tested 10 times. 

\textbf{Clutter Pick-place Task.} In this task, the robot cannot directly grasp the object, as it is tightly surrounded by distractors. Therefore, it must first perform push actions to create sufficient clearance for gripper insertion before executing the pick-place actions~\cite{xu2021efficient,chen2025clutterdexgrasp}. This requires long-horizon planning across push, pick, and place sub-tasks. Additionally, the task demands identifying the target object among multiple objects and selecting the correct container among several options, emphasizing semantic grounding. The language prompt for this task is: “Pick up the \{object\} and place it on the \{container\},” where the \{object\} denotes the target object to pick, and \{container\} specifies the placement target. Representative successful rollout and example cases of different generalization dimensions are shown in Figure~\ref{fig:real-bench}. Each policy is evaluated over 80 trials in total. In the first 30 trials, the target objects and distractors are seen during training. For the rest 50 trials, we test the generalization of: seen object with unseen container~(UC), unseen object with seen container~(UO), and unseen object with unseen background~(UOUE). Each configuration is tested 10 times. 

\subsection{Compositional Task Detailed Settings}
\label{subsec:compositional_appendix}
\begin{table}[t]
\centering
\begin{tabular}{ccccc}
\toprule
Method & Red Tea Bag & Spoon & Box & Avg. \\
\midrule
$\pi_{0.5}$ & 19/20 & 19/20 & 16/20 & 16/20 \\ 
\ours  & 20/20 & 18/20 & 17/20 & 17/20 \\
\bottomrule
\end{tabular}
\vspace{0.2cm}
\caption{Real-world table storage results (successes / trials).}
\vspace{-0.5cm}
\label{tab:real-storage}
\end{table}

We evaluate compositional instruction following and generalization across two real-world tasks:
\begin{itemize}[leftmargin=*, topsep=2pt, itemsep=1pt]
    \item \textbf{Task 1: Table Storage (T1).} The robot picks specified items from the table, places them sequentially in a storage box, and finally pushes the box closed. The training instruction template is ``From the table, put the \{objects\} into the storage box. Then close the box.'', where \{objects\} are drawn from a fixed set of seen items including a marker, a spoon, a tape, and a red tea bag.
    \item \textbf{Task 2: Pick-Place (T2).} Same as the single pick-place task in the previous subsection, using only seen objects and seen containers.
\end{itemize}
We first test \textbf{Task 1} of seen prompt ``From the table, put the red tea bag, and the spoon into the storage box. Finally, close the box.'' Combining results in Table~\ref{tab:real-storage} and Table~\ref{tab:real-task-gen}~(Pick-Place SO), we can confirm that the two policies have sufficient single-task competence. For the compositional evaluation, all objects are drawn from the seen training set; only the instruction structure or wording, and the object layouts are varied.

We design three evaluation protocols for 2 compositional tasks of increasing difficulty:

\textbf{Seen Language Coaching.} Instructions for the two tasks are issued sequentially as separate prompts using the training-distribution wording. For example, the user first prompts T1 with ``From the table, put the red tea bag into the storage box. Then close the box.'', and after completion, prompts T2 with ``Pick up the pepper and place it on the blue plate.'' We test 20 trials in total, 10 with T1 issued before T2, and 10 with T2 issued before T1.
 
\textbf{Unseen Language Coaching.} The same two-prompt protocol, but each prompt is rewritten along one of three OOD axes:
(i) \textit{Verb substitution}, replacing the training verb while keeping the task semantics, \textit{e.g.}, ``Place the red tea bag into the box. Then push the storage box closed.'';
(ii) \textit{Noun substitution}, replacing the object or container noun with a synonym, \textit{e.g.}, ``Put the red tea pouch into the bin. Then close the bin.'';
(iii) \textit{Task substitution}, swapping the manipulated objects between the two tasks, \textit{e.g.}, T1 becomes ``Put the pepper into the box. Then close the storage box.'' and T2 becomes ``Pick up the red tea bag and place it on the blue plate.'' These perturbations probe the policy's ability to compose seen objects into novel task assignments. 
 
\textbf{Task Chaining.} The two tasks are combined into a single concatenated prompt, for instance ``Put the red tea bag into the box. Then close the storage box. Pick up the pepper and place it on the blue plate.''. The policy must parse and execute the multi-task instruction without an explicit segmentation signal.




\subsection{Detailed Results}

\textbf{Pick-Place Task.} Table~\ref{tab:real-pp-detailed} shows per-object results on the single pick-place task. In the SO split, both methods reliably handle the carrot and eggplant. The gap widens on unseen objects: in UO, $\pi_{0.5}$ achieves only $5/10$ and $6/10$ on grape and bottle, whereas \ours reaches $9/10$ and $8/10$. Under UOUC and UOUCUE, $\pi_{0.5}$'s success rates further degrade, particularly on the bottle ($6/20$ in UOUCUE), where its narrow shape under unseen backgrounds frequently leads to misgrounding or misgrasping. \ours maintains a clear margin across all splits~($15/20$ and $13/20$ on grape and bottle in UOUCUE), indicating that the pretrained VA prior together with gated language fusion is robust to compound shifts in object identity, container, and environment.

\textbf{Clutter Pick-Place Task.} Table~\ref{tab:real-clutter-detailed} shows per-object results on the clutter pick-place task. \ours outperforms $\pi_{0.5}$ on almost every object and every split, with the largest gap on the seen-object pepper case in SO ($9/10$ vs.\ $4/10$) and on the unseen-object grape under unseen backgrounds in UOUE ($6/10$ vs.\ $3/10$). The pepper case is notable: although the object is seen, $\pi_{0.5}$ often hesitates during the push-to-grasp transition when the pepper shares a similar height with surrounding distractors, a behavior that the pretrained VA prior largely avoids. Across UC, UO, and UOUE, $\pi_{0.5}$ is most prone to misgrounding when the target shares color with distractors~(grape vs.\ eggplant), while \ours maintains correct grounding and produces a coherent push-pick-place trajectory.

\textbf{Statistical Significance.} Table~\ref{tab:real_stats_appendix} reports two-sided Fisher's exact tests of APT and $\pi_{0.5}$. APT shows statistically significant improvements on Pick-Place, Clutter Pick-Place, and task chaining~($p<0.05$).

\begin{table}[t]
    \centering
    \footnotesize
    \setlength{\tabcolsep}{4pt}
    \begin{tabular}{l ccc cc cc cc c}
        \toprule
        \multirow{2}{*}{Method}
            & \multicolumn{3}{c}{SO}
            & \multicolumn{2}{c}{UO}
            & \multicolumn{2}{c}{UOUC}
            & \multicolumn{2}{c}{UOUCUE}
            & \multirow{2}{*}{Avg.} \\
        \cmidrule(lr){2-4}\cmidrule(lr){5-6}\cmidrule(lr){7-8}\cmidrule(lr){9-10}
            & Pepper & Carrot & Eggplant
            & Grape & Bottle
            & Grape & Bottle
            & Grape & Bottle
            & \\
        \midrule
        $\pi_{0.5}$~\cite{black2025pi_ohfive}
            & 7/10 & \textbf{10/10} & \textbf{10/10}
            & 5/10 & 6/10
            & 4/10 & 5/10
            & 10/20 & 6/20
            & 63/110 \\
        \textbf{\ours}
            & \textbf{9/10} & \textbf{10/10} & \textbf{10/10}
            & \textbf{9/10} & \textbf{8/10}
            & \textbf{9/10} & \textbf{7/10}
            & \textbf{15/20} & \textbf{13/20}
            & \textbf{90/110} \\
        \bottomrule
    \end{tabular}
    \vspace{0.2cm}
    \caption{Real-world pick-place per-object results (successes / trials).}
    \vspace{-0.5cm}
    \label{tab:real-pp-detailed}
\end{table}

\begin{table}[t]
    \centering
    \footnotesize
    \setlength{\tabcolsep}{4pt}
    \begin{tabular}{l ccc ccc c c c}
        \toprule
        \multirow{2}{*}{Method}
            & \multicolumn{3}{c}{SO}
            & \multicolumn{3}{c}{UC}
            & UO & UOUE
            & \multirow{2}{*}{Avg.} \\
        \cmidrule(lr){2-4}\cmidrule(lr){5-7}\cmidrule(lr){8-8}\cmidrule(lr){9-9}
            & Pepper & Carrot & Eggplant
            & Pepper & Carrot & Eggplant
            & Grape & Grape
            & \\
        \midrule
        $\pi_{0.5}$~\cite{black2025pi_ohfive}
            & 4/10 & 7/10 & 7/10
            & 4/10 & \textbf{8/10} & 6/10
            & 4/10 & 3/10
            & 43/80 \\
        \textbf{\ours}
            & \textbf{9/10} & \textbf{8/10} & \textbf{8/10}
            & \textbf{7/10} & \textbf{8/10} & \textbf{7/10}
            & \textbf{7/10} & \textbf{6/10}
            & \textbf{60/80} \\
        \bottomrule
    \end{tabular}
    \vspace{0.2cm}
    \caption{Real-world clutter pick-place per-object results (successes / trials).}
    \vspace{-0.5cm}
    \label{tab:real-clutter-detailed}
\end{table}

\begin{table}[t]
    \centering
      \begin{tabular}{cccc}
        \toprule
        Task & {Pick-Place} & {Clutter Pick-Place} & {Composition} \\
        \midrule
        \multirow{2}{*}{Split-level}
          & SO 0.612; UO 0.082
          & SO 0.084; UC 0.412
          & Coaching 0.108 \\
          & UOUC {0.048}; UOUCUE {0.013}
          & UO 0.370; UOUE 0.370
          & Chaining {0.000013} \\
        Overall & {0.00012} & {0.008} & -- \\
        \bottomrule
      \end{tabular}
    \vspace{0.2cm}
    \caption{Real-world two-sided Fisher's exact-test $p$-values. }
    \vspace{-0.5cm}
    \label{tab:real_stats_appendix}
\end{table}

\subsection{Case Studies}
\label{subsec:real_cases}

\begin{figure}[t]
    \centering
    \includegraphics[width=\linewidth]{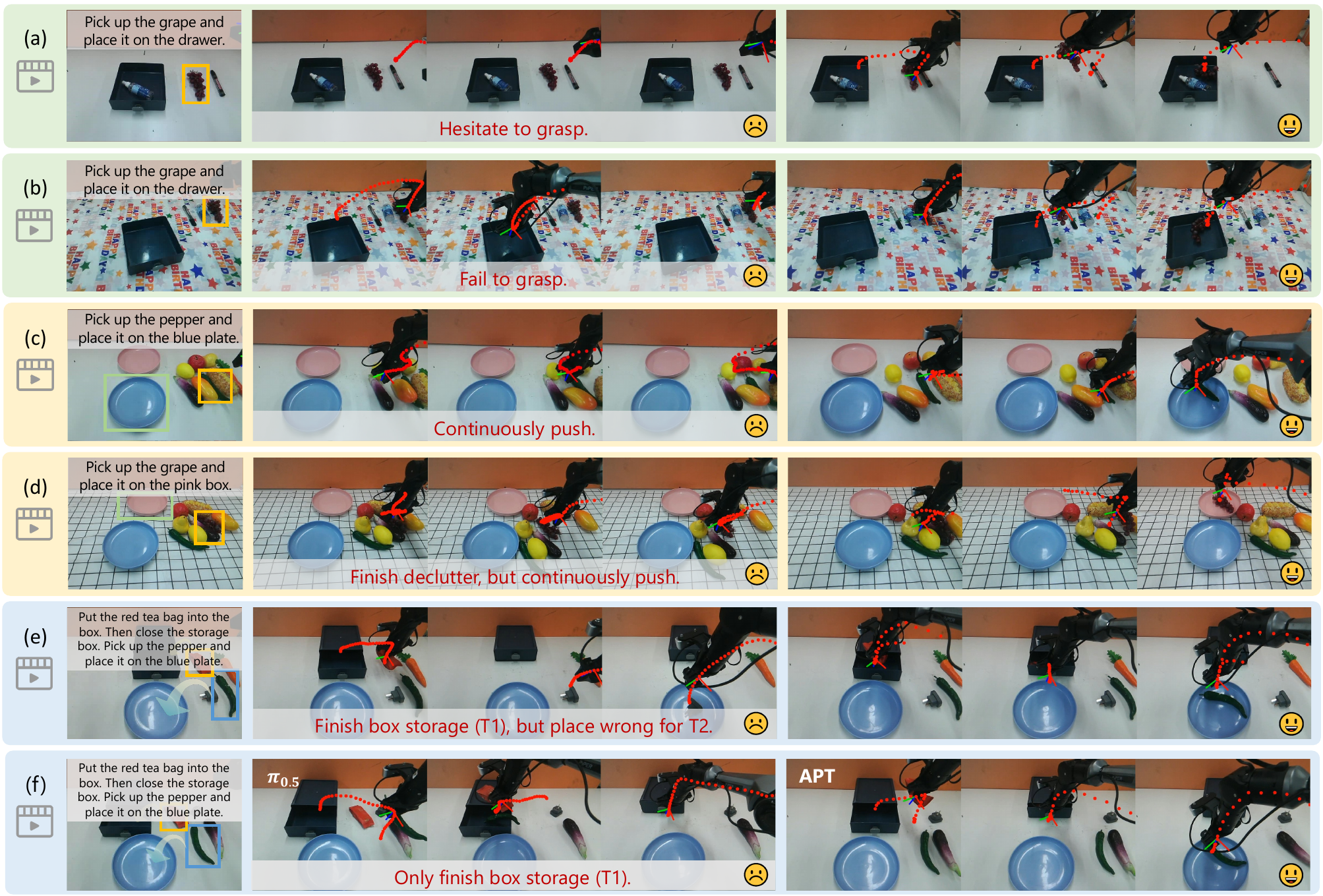}
    \caption{
        More real-world cases comparing $\pi_{0.5}$ and \ours. (a, b) Grasping on unseen objects. (c, d) Sub-task transition in clutter pick-place. (e, f) Compositional task execution. Red dotted lines visualize end-effector trajectories; annotated text highlights the failure cause for $\pi_{0.5}$.
    }
    \label{fig:real-cases-more}
\end{figure}

Figure~\ref{fig:real-cases-more} presents additional real-world cases across the single-task and compositional-task settings.

\textbf{Grasping on Unseen Objects~(a, b).} On UOUC and UOUCUE pick-place cases with the unseen grape, $\pi_{0.5}$ either hesitates without closing the gripper or slips off on contact, while \ours grasps the target on the first attempt. We also observe that when a non-target object is initially placed inside the container~(a), $\pi_{0.5}$ often remains completely still, likely misinterpreting the task as already completed. 
 
\textbf{Sub-task Transition in Clutter~(c, d).} The clutter pick-place task requires a push-grasp-place sequence, and $\pi_{0.5}$ fails in two distinct modes at this transition. In (c), the target pepper is still partially surrounded by distractors and $\pi_{0.5}$ keeps issuing push actions without creating enough space for grasping. In (d), the policy successfully declutters and the target grape is fully exposed, yet it continues to push rather than switching to grasp, suggesting that the sub-task boundary is missed even when the perceptual condition for grasping is clearly met. In contrast, \ours transitions from push to grasp at the appropriate time in both cases. These failures show that the bottleneck is not perceptual access to the target, but the policy's ability to commit to the next sub-task once the current one's precondition is satisfied, which is harder to learn when the action expert is jointly trained from random initialization rather than pretrained as a structured prior.
 
\textbf{Compositional Execution~(e, f).} On chained instructions combining T1~(storage) and T2~(pick-place), $\pi_{0.5}$ fails in two distinct modes. In (e), it completes T1 correctly but during T2 places the wrong object into the plate, indicating that the chained instruction affects its language grounding. In (f), it executes T1, but places both the T1 and T2 target objects into the box and then stops. This suggests that the concatenated prompt is treated as a single task whose completion is judged by the T1 sub-goal alone. In contrast, \ours executes both sub-tasks in the correct order and with the correct targets. These two failure modes show that the gap is not about understanding individual instructions, since each sub-task is reliable in isolation, but about parsing and switching between sub-instructions within a single prompt. This is consistent with the compositional-task results in Figure~\ref{fig:compositional} and indicates that chained multi-task prompts are where the language-grounding gap is most visible.

\subsection{Failure Analysis}

\begin{figure}[t]
    \centering
    \includegraphics[width=0.8\linewidth]{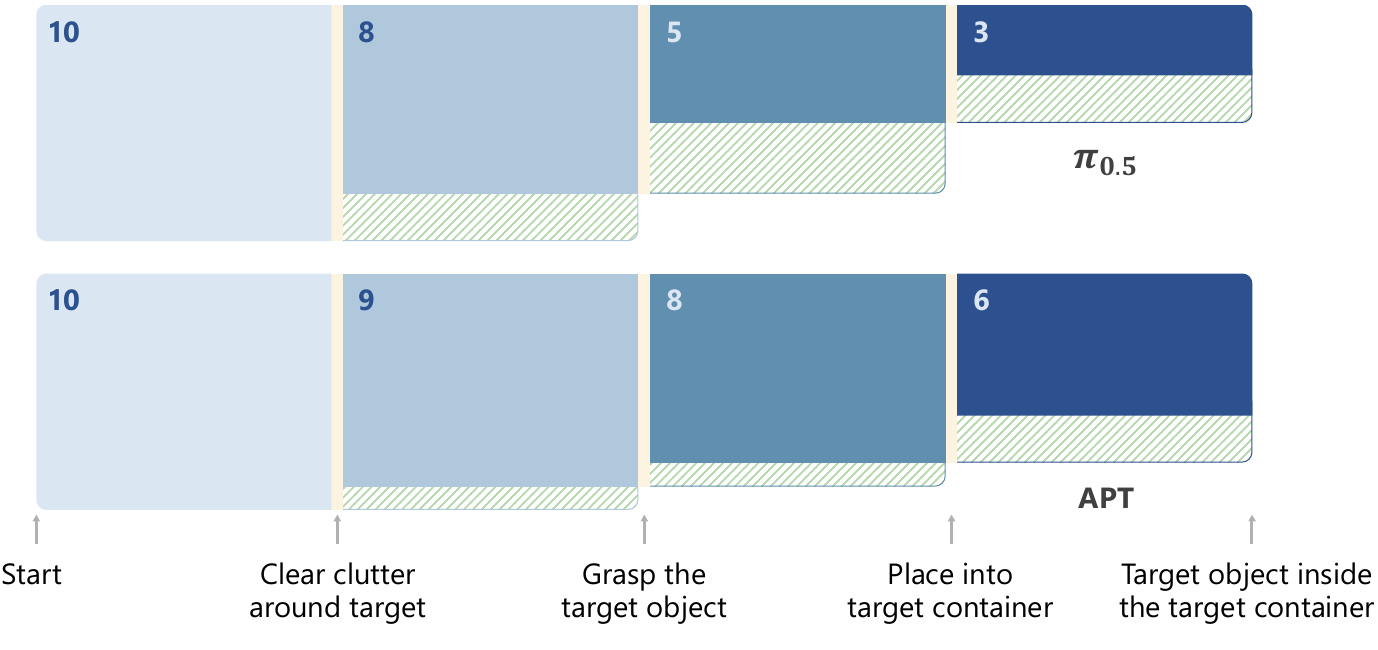}
    \caption{
        Failure breakdown on the UOUE setting of the clutter pick-place task. The Sankey diagram tracks the flow of success (solid) and failure (shadow) across the start, clear clutter, grasp, and place sub-tasks.
    }
    \label{fig:failure-analysis}
\end{figure}

\begin{figure}[t]
    \centering
    \includegraphics[width=0.6\linewidth]{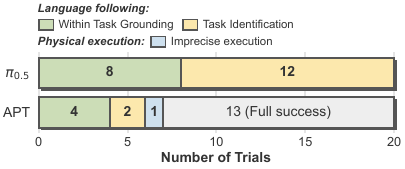}
    \caption{Failure categories and full successes across 20 task-chaining trials per policy. Language-related failures comprise task-identification and within-task grounding errors; physical-execution failures are shown separately.}
    \label{fig:chaining_failure_appendix}
\end{figure}

\textbf{Failure Analysis.} Figure~\ref{fig:failure-analysis} visualizes the failure breakdown on the UOUE setting of the clutter pick-place tasks. The four checkpoints are: clear clutter around the target, grasp the target object, place into the target container, and target object inside the target container. Starting from 10 trials, $\pi_{0.5}$ retains $10 \to 8 \to 5 \to 3 \to 3$ rollouts across the four checkpoints, while \ours retains $10 \to 9 \to 8 \to 6 \to 6$. The largest gap appears at the push-to-grasp transition. Common causes for $\pi_{0.5}$ are: (i) confusing the grape with an eggplant when their colors overlap, and (ii) continuing to push after the target is already exposed, both of which point to weak target-object grounding under unseen visual conditions. \ours retains correct grounding in most rollouts; its remaining failures are concentrated at the contact and placement stages, which are physical execution issues rather than language-following issues. Figure~\ref{fig:chaining_failure_appendix} categorizes the 20 task-chaining trials per policy into full success, language-following failure, and physical-execution failure. Language-following failures are further divided into task-identification errors, such as executing the wrong subtask or switching at the wrong time, and within-task grounding errors involving the instructed target. All 20 failures of $\pi_{0.5}$ are language-related: 12 involve task identification and 8 involve within-task grounding. APT has 2 task-identification errors, 4 grounding errors, and 1 physical-execution error, with 13 full successes. Thus, in this evaluation, the performance gain is primarily associated with fewer language-following failures, especially fewer task-identification errors. 

\begin{figure}[t]
    \centering
    \includegraphics[width=0.8\linewidth]{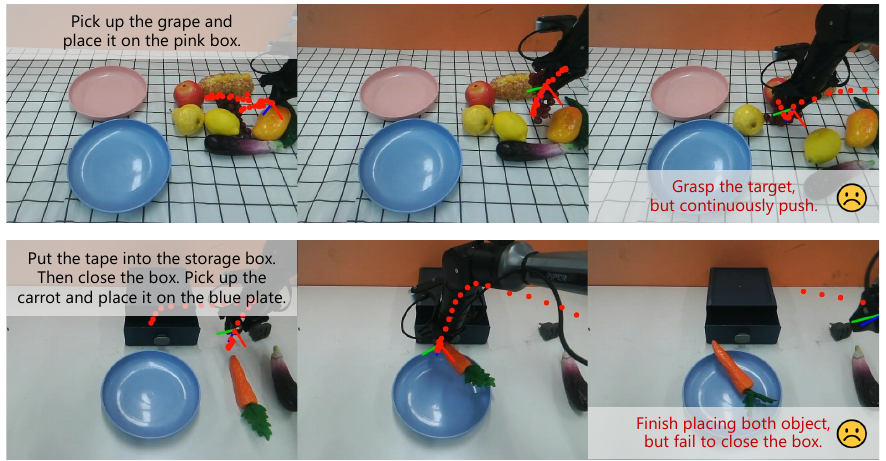}
    \caption{Typical failure cases of \ours. Top: continued pushing after successfully grasping the target on clutter pick-place. Bottom: skipped ``close the box'' action at the end of T1 on compositional task chaining.}
    \label{fig:failure-cases}
\end{figure}

\textbf{Failure Cases of \ours.} We show the typical failure modes of \ours in Figure~\ref{fig:failure-cases}. On clutter pick-place, \ours occasionally exhibits the same push-after-grasp behavior described earlier for $\pi_{0.5}$, where the policy successfully grasps the target but continues to issue push actions instead of transitioning to placement. The frequency of this failure is much lower than for $\pi_{0.5}$, but it has not been fully eliminated. On compositional task chaining, \ours occasionally over-attends to the pick-place sub-task and skips the ``close the box'' action at the end of T1, completing the placement portion of the chained prompt while leaving the box open. Both failures indicate a remaining gap for \ours on detecting sub-task termination.

\section{Additional Ablations and Analysis}
\label{sec:additional_analyses}

\subsection{Attention Layer-Split Sensitivity}
\label{subsec:layer_allocation}

Table~\ref{tab:layer_split_appendix} ablates the fraction of attention layers pretrained in Stage 1 versus those newly introduced in Stage 2. The 100/0 configuration pretrains all layers and adds no interleaved layers to test the contribution of inserting additional layers. Among the evaluated splits, 50/50 gives the highest success rates across all four suites. This result supports allocating comparable capacity to VA-prior learning and VLA alignment.

\begin{table}[t]
    \centering
    \begin{tabular}{lcccc}
        \toprule
        Layer Split & SO & UO & UC & UOUE \\
        \midrule
        50/50 & \textbf{96} & \textbf{74} & \textbf{90} & \textbf{62} \\
        80/20 & 90 & 64 & 68 & 58 \\
        100/0 & 92 & 56 & 70 & 44 \\
        \bottomrule
    \end{tabular}
    \vspace{0.2cm}
    \caption{Layer-split sensitivity on rigid object pick-place (success rate \%). The split specifies the fractions of layers pretrained in Stage 1 and newly introduced in Stage 2.}
    \vspace{-0.5cm}
    \label{tab:layer_split_appendix}
\end{table}

\subsection{Multimodal Action Distributions under Underspecified Instructions}
\label{subsec:multimodal_prior}

\begin{figure}[t]
    \centering
    \includegraphics[width=0.8\linewidth]{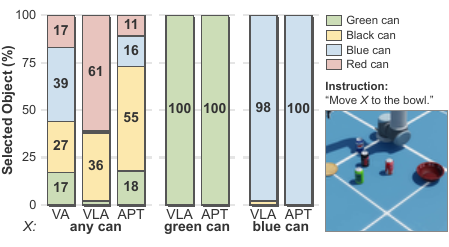}
    \caption{Object-selection frequencies under an underspecified instruction and target-specific instructions in the same scene. VA denotes the language-agnostic Stage 1 policy, VLA denotes one-stage language-conditioned training, and APT denotes the two-stage policy.}
    \label{fig:multimodal_prior_appendix}
\end{figure}

In APT, Stage 1 learns a vision-conditioned action distribution~(VA) with diffusion. When several objects can be manipulated in the same scene, this distribution can represent multiple plausible actions, rather than forcing a deterministic regressor to average incompatible actions. Language conditioning in Stage 2 can then select among these possibilities. To validate this capability, Figure~\ref{fig:multimodal_prior_appendix} compares VA, one-stage VLA, and two-stage VLA~(APT) in a scene with four cans. Under the unseen underspecified instruction \emph{Move any can to the bowl}, VA and APT select all four objects with nonzero frequencies, while one-stage VLA concentrates 61\% of selections on the red can and 36\% on the black can. With target-specific instructions, both language-conditioned VLA models shift toward the specified object. This demonstrates that APT can retain multiple action modes while enabling instruction-dependent selection.

\subsection{VLM Feature Analysis and Policy Language Following}
\label{subsec:language_following_analysis}

\begin{figure}[t]
    \centering
    \includegraphics[width=\linewidth]{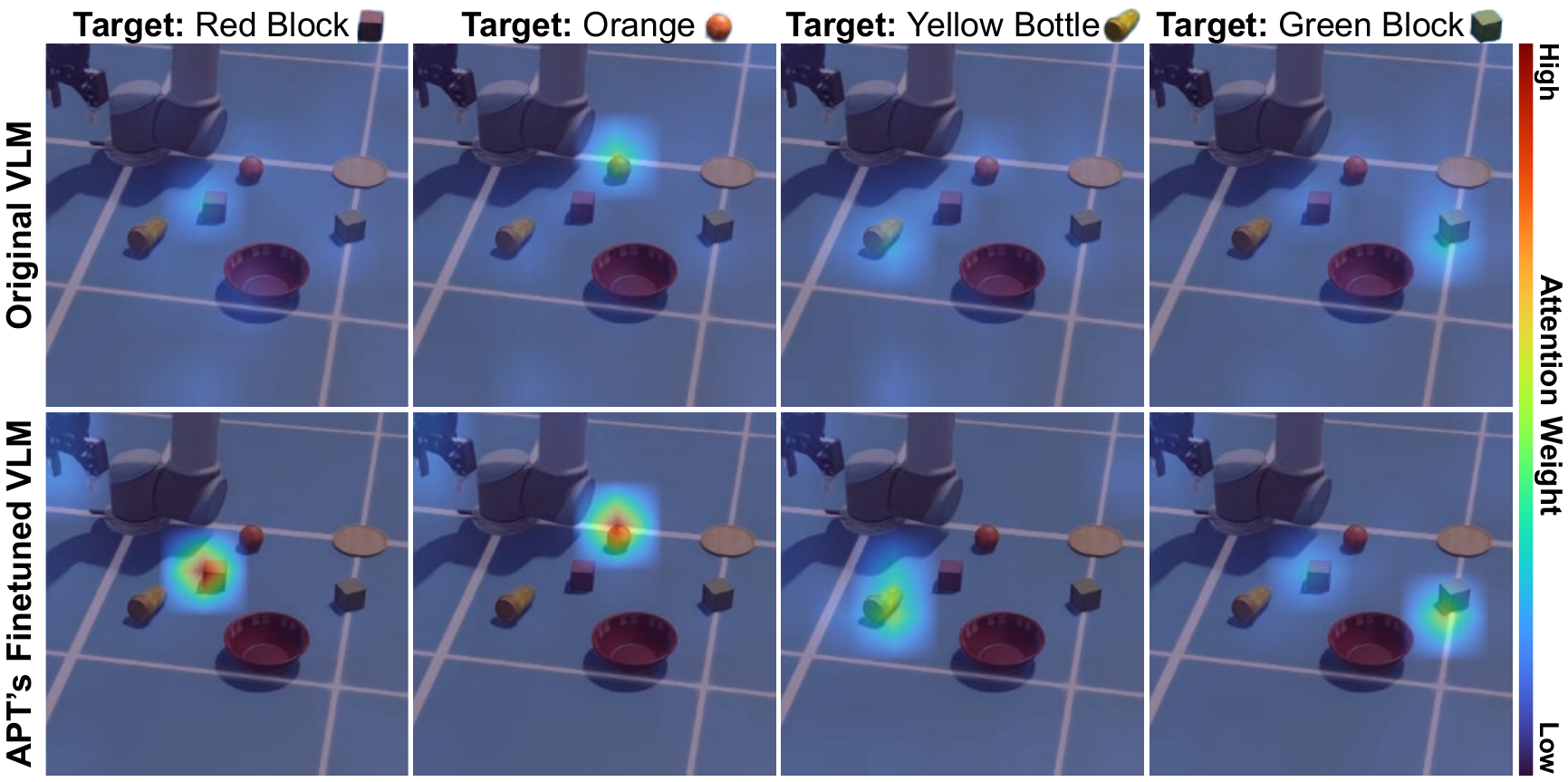}
    \caption{Vision-language attention with a fixed observation and different target instructions. Top: original VLM. Bottom: APT's finetuned VLM.}
    \label{fig:vl_attention_appendix}
\end{figure}

\textbf{VLM Attention.} We hold the observation fixed and vary the instructed target among four objects in the scene. Figure~\ref{fig:vl_attention_appendix} compares the attention maps of vision-language features between the original VLM and APT's finetuned VLM. The finetuned VLM exhibits more concentrated attention around the instructed targets in these examples. This provides qualitative evidence of instruction-dependent alignment.

\textbf{Language Following Independently of Execution.} On rigid object pick-place, we also evaluate language following separately from physical execution success. Table~\ref{tab:language_following_appendix} reports these rates on the SO, UO, UC, and UOUE suites. APT matches $\pi_{0.5}$ on SO and improves on the three generalization suites, with the largest difference on UOUE (82\% versus 62\%). This distinction helps separate instruction-grounding errors from failures in executing an otherwise correctly targeted action.

\begin{table}[t]
    \centering
    \begin{tabular}{lcccc}
        \toprule
        Method & SO & UO & UC & UOUE \\
        \midrule
        $\pi_{0.5}$ & 98 & 82 & 94 & 62 \\
        \ours & 98 & \textbf{88} & \textbf{96} & \textbf{82} \\
        \bottomrule
    \end{tabular}
    \vspace{0.2cm}
    \caption{Language-following rates (\%) on rigid object pick-place, evaluated independently of physical execution success. These rates are distinct from task success rates in Table~\ref{tab:pickplace_results}.}
    \label{tab:language_following_appendix}
\end{table}

%% file: ref.bib
@inproceedings{kim2024openvla,
  title={OpenVLA: An open-source vision-language-action model},
  author={Kim, Moo Jin and Pertsch, Karl and Karamcheti, Siddharth and Xiao, Ted and Balakrishna, Ashwin and Nair, Suraj and Rafailov, Rafael and Foster, Ethan P and Sanketi, Pannag R and Vuong, Quan and others},
  booktitle={Conference on Robot Learning (CoRL)},
  pages={2679--2713},
  year={2024}
}

@inproceedings{liu2024rdt,
  title     = {{RDT-1B}: A diffusion foundation model for bimanual manipulation},
  author    = {Liu, Songming and Wu, Lingxuan and Li, Bangguo and Tan, Hengkai and Chen, Huayu and Wang, Zhengyi and Xu, Ke and Su, Hang and Zhu, Jun},
  booktitle = {International Conference on Learning Representations (ICLR)},
  year      = {2025}
}

@article{brohan2022rt1,
  title     = {{RT-1}: Robotics transformer for real-world control at scale},
  author    = {Brohan, Anthony and Brown, Noah and Carbajal, Justice and Chebotar, Yevgen and Dabis, Joseph and Finn, Chelsea and Gopalakrishnan, Keerthana and Hausman, Karol and Herzog, Alexander and Hsu, Jasmine and others},
  journal   = {Robotics: Science and Systems (RSS)},
  year      = {2023},
  publisher = {Robotics: Science and Systems Foundation}
}

@article{black2024pi0,
  title={$\pi_0 $: A vision-language-action flow model for general robot control},
  author={Black, Kevin and Brown, Noah and Driess, Danny and Esmail, Adnan and Equi, Michael and Finn, Chelsea and Fusai, Niccolo and Groom, Lachy and Hausman, Karol and Ichter, Brian and others},
  journal={arXiv preprint arXiv:2410.24164},
  year={2024}
}

@article{black2025pi_ohfive,
  title={$\pi_{0.5}$: A vision-language-action model with open-world generalization},
  author={Intelligence, Physical and Black, Kevin and Brown, Noah and Darpinian, James and Dhabalia, Karan and Driess, Danny and Esmail, Adnan and Equi, Michael and Finn, Chelsea and Fusai, Niccolo and others},
  journal={arXiv preprint arXiv:2504.16054},
  year={2025}
}

@article{beyer2024paligemma,
  title={Paligemma: A versatile 3b vlm for transfer},
  author={Beyer, Lucas and Steiner, Andreas and Pinto, Andr{\'e} Susano and Kolesnikov, Alexander and Wang, Xiao and Salz, Daniel and Neumann, Maxim and Alabdulmohsin, Ibrahim and Tschannen, Michael and Bugliarello, Emanuele and others},
  journal={arXiv preprint arXiv:2407.07726},
  year={2024}
}

@misc{openxembodiment,
title={Open {X-E}mbodiment: Robotic learning datasets and {RT-X} models},
author = {Open X-Embodiment Collaboration},
howpublished  = {\url{https://arxiv.org/abs/2310.08864}},
year = {2023},
}

@inproceedings{khazatsky2024droid,
  title     = {{DROID}: A large-scale in-the-wild robot manipulation dataset},
  author    = {Khazatsky, Alexander and Pertsch, Karl and Nair, Suraj and Balakrishna, Ashwin and Dasari, Sudeep and Karamcheti, Siddharth and Nasiriany, Soroush and Srirama, Mohan Kumar and Chen, Lawrence Yunliang and Ellis, Kirsty and others},
  booktitle = {Robotics: Science and Systems (RSS)},
  year      = {2024}
}

@article{kim2025oft,
  title={Fine-tuning vision-language-action models: Optimizing speed and success},
  author={Kim, Moo Jin and Finn, Chelsea and Liang, Percy},
  journal={arXiv preprint arXiv:2502.19645},
  year={2025}
}

@article{pertsch2025fast,
  title={Fast: Efficient action tokenization for vision-language-action models},
  author={Pertsch, Karl and Stachowicz, Kyle and Ichter, Brian and Driess, Danny and Nair, Suraj and Vuong, Quan and Mees, Oier and Finn, Chelsea and Levine, Sergey},
  journal={arXiv preprint arXiv:2501.09747},
  year={2025}
}

@article{ho2020ddpm,
  title={Denoising diffusion probabilistic models},
  author={Ho, Jonathan and Jain, Ajay and Abbeel, Pieter},
  journal={Advances in Neural Information Processing Systems (NeurIPS)},
  volume={33},
  pages={6840--6851},
  year={2020}
}

@inproceedings{wen2025diffusionvla,
  title={DiffusionVLA: Scaling robot foundation models via unified diffusion and autoregression},
  author={Wen, Junjie and Zhu, Yichen and Zhu, Minjie and Tang, Zhibin and Li, Jinming and Zhou, Zhongyi and Liu, Xiaoyu and Shen, Chaomin and Peng, Yaxin and Feng, Feifei},
  booktitle={International Conference on Machine Learning (ICML)},
  pages={66558--66574},
  year={2025},
  organization={PMLR}
}

@article{team2024octo,
  title={Octo: An open-source generalist robot policy},
  author={Team, Octo Model and Ghosh, Dibya and Walke, Homer and Pertsch, Karl and Black, Kevin and Mees, Oier and Dasari, Sudeep and Hejna, Joey and Kreiman, Tobias and Xu, Charles and others},
  journal={arXiv preprint arXiv:2405.12213},
  year={2024}
}

@inproceedings{zitkovich2023rt2,
  title={RT-2: Vision-language-action models transfer web knowledge to robotic control},
  author={Zitkovich, Brianna and Yu, Tianhe and Xu, Sichun and Xu, Peng and Xiao, Ted and Xia, Fei and Wu, Jialin and Wohlhart, Paul and Welker, Stefan and Wahid, Ayzaan and others},
  booktitle={Conference on Robot Learning (CoRL)},
  pages={2165--2183},
  year={2023},
  organization={PMLR}
}

@article{shukor2025smolvla,
  title={Smolvla: A vision-language-action model for affordable and efficient robotics},
  author={Shukor, Mustafa and Aubakirova, Dana and Capuano, Francesco and Kooijmans, Pepijn and Palma, Steven and Zouitine, Adil and Aractingi, Michel and Pascal, Caroline and Russi, Martino and Marafioti, Andres and others},
  journal={arXiv preprint arXiv:2506.01844},
  year={2025}
}

@article{bjorck2025gr00tn1,
  title={Gr00t n1: An open foundation model for generalist humanoid robots},
  author={Bjorck, Johan and Casta{\~n}eda, Fernando and Cherniadev, Nikita and Da, Xingye and Ding, Runyu and Fan, Linxi and Fang, Yu and Fox, Dieter and Hu, Fengyuan and Huang, Spencer and others},
  journal={arXiv preprint arXiv:2503.14734},
  year={2025}
}

@article{bjorck2025gr00tn15,
  title={GR00T N1.5: An improved open foundation model for generalist humanoid robots},
  author={Bjorck, Johan and Casta{\~n}eda, Fernando and Cherniadev, Nikita and Da, Xingye and Ding, Runyu and Fan, Linxi and Fang, Yu and Fox, Dieter and Hu, Fengyuan and Huang, Spencer and others},
  year={2025}
}

@inproceedings{chi2023dp,
    title={Diffusion policy: Visuomotor policy learning via action diffusion},
    author={Chi, Cheng and Feng, Siyuan and Du, Yilun and Xu, Zhenjia and Cousineau, Eric and Burchfiel, Benjamin and Song, Shuran},
    booktitle={Robotics: Science and Systems (RSS)},
    year={2023}
}

@article{liu2023libero,
  title={Libero: Benchmarking knowledge transfer for lifelong robot learning},
  author={Liu, Bo and Zhu, Yifeng and Gao, Chongkai and Feng, Yihao and Liu, Qiang and Zhu, Yuke and Stone, Peter},
  journal={Advances in Neural Information Processing Systems (NeurIPS)},
  volume={36},
  pages={44776--44791},
  year={2023}
}

@article{gao2025taxonomy,
  title={A taxonomy for evaluating generalist robot policies},
  author={Gao, Jensen and Belkhale, Suneel and Dasari, Sudeep and Balakrishna, Ashwin and Shah, Dhruv and Sadigh, Dorsa},
  journal={arXiv preprint arXiv:2503.01238},
  year={2025}
}

@inproceedings{driess2025knowledge,
  title     = {Knowledge insulating vision-language-action models: Train fast, run fast, generalize better},
  author    = {Driess, Danny and Springenberg, Jost Tobias and Yu, Lili and Li-Bell, Adrian and Pertsch, Karl and Ren, Allen Z and Walke, Homer and Vuong, Quan and Shi, Lucy Xiaoyang and Levine, Sergey and others},
  booktitle = {Advances in Neural Information Processing Systems (NeurIPS)},
  year      = {2025}
}

@article{zhou2025liberopro,
  title={LIBERO-PRO: Towards robust and fair evaluation of vision-language-action models beyond memorization},
  author={Zhou, Xueyang and Xu, Yangming and Tie, Guiyao and Chen, Yongchao and Zhang, Guowen and Chu, Duanfeng and Zhou, Pan and Sun, Lichao},
  journal={arXiv preprint arXiv:2510.03827},
  year={2025}
}

@article{chen2025e2vla,
  title={Toward embodiment equivariant vision-language-action policy},
  author={Chen, Anzhe and Yang, Yifei and Zhu, Zhenjie and Xu, Kechun and Zhou, Zhongxiang and Xiong, Rong and Wang, Yue},
  journal={arXiv preprint arXiv:2509.14630},
  year={2025}
}

@INPROCEEDINGS{zhou2019rot6d,
  author={Zhou, Yi and Barnes, Connelly and Lu, Jingwan and Yang, Jimei and Li, Hao},
  booktitle={Proceedings of the IEEE/CVF Conference on Computer Vision and Pattern Recognition (CVPR)}, 
  title={On the continuity of rotation representations in neural networks}, 
  year={2019},
  volume={},
  number={},
  pages={5738-5746},
}

@article{su2024roformer,
  title={Roformer: Enhanced transformer with rotary position embedding},
  author={Su, Jianlin and Ahmed, Murtadha and Lu, Yu and Pan, Shengfeng and Bo, Wen and Liu, Yunfeng},
  journal={Neurocomputing},
  volume={568},
  pages={127063},
  year={2024},
  publisher={Elsevier}
}

@inproceedings{jiang2023vima,
  title={VIMA: Robot manipulation with multimodal prompts},
  author={Jiang, Yunfan and Gupta, Agrim and Zhang, Zichen and Wang, Guanzhi and Dou, Yongqiang and Chen, Yanjun and Fei-Fei, Li and Anandkumar, Anima and Zhu, Yuke and Fan, Linxi},
  booktitle={International Conference on Machine Learning (ICML)},
  pages={14975--15022},
  year={2023}
}

@inproceedings{perez2018film,
  title={Film: Visual reasoning with a general conditioning layer},
  author={Perez, Ethan and Strub, Florian and De Vries, Harm and Dumoulin, Vincent and Courville, Aaron},
  booktitle={Proceedings of the AAAI Conference on Artificial Intelligence (AAAI)},
  volume={32},
  number={1},
  pages={3942--3951},
  year={2018}
}

@article{cui2025openhelix,
  title={Openhelix: A short survey, empirical analysis, and open-source dual-system vla model for robotic manipulation},
  author={Cui, Can and Ding, Pengxiang and Song, Wenxuan and Bai, Shuanghao and Tong, Xinyang and Ge, Zirui and Suo, Runze and Zhou, Wanqi and Liu, Yang and Jia, Bofang and others},
  journal={arXiv preprint arXiv:2505.03912},
  year={2025}
}

@inproceedings{karamcheti2024prismatic,
  title={Prismatic vlms: Investigating the design space of visually-conditioned language models},
  author={Karamcheti, Siddharth and Nair, Suraj and Balakrishna, Ashwin and Liang, Percy and Kollar, Thomas and Sadigh, Dorsa},
  booktitle={International Conference on Machine Learning (ICML)},
  pages={23123--23144},
  year={2024}
}

@software{NVIDIA_Isaac_Sim,
author = {{NVIDIA}},
license = {Apache-2.0},
title = {{Isaac Sim}},
url = {https://github.com/isaac-sim/IsaacSim},
version = {5.1.0}
}

@inproceedings{fang2024rh20t,
  title={Rh20t: A comprehensive robotic dataset for learning diverse skills in one-shot},
  author={Fang, Hao-Shu and Fang, Hongjie and Tang, Zhenyu and Liu, Jirong and Wang, Chenxi and Wang, Junbo and Zhu, Haoyi and Lu, Cewu},
  booktitle={IEEE International Conference on Robotics and Automation (ICRA)},
  pages={653--660},
  year={2024},
  organization={IEEE}
}

@article{wu2024robomind,
  title={Robomind: Benchmark on multi-embodiment intelligence normative data for robot manipulation},
  author={Wu, Kun and Hou, Chengkai and Liu, Jiaming and Che, Zhengping and Ju, Xiaozhu and Yang, Zhuqin and Li, Meng and Zhao, Yinuo and Xu, Zhiyuan and Yang, Guang and others},
  journal={arXiv preprint arXiv:2412.13877},
  year={2024}
}

@article{cheang2025gr3,
  title={Gr-3 technical report},
  author={Cheang, Chilam and Chen, Sijin and Cui, Zhongren and Hu, Yingdong and Huang, Liqun and Kong, Tao and Li, Hang and Li, Yifeng and Liu, Yuxiao and Ma, Xiao and others},
  journal={arXiv preprint arXiv:2507.15493},
  year={2025}
}

@article{bu2025agibotworld,
  title={Agibot world colosseo: A large-scale manipulation platform for scalable and intelligent embodied systems},
  author={Bu, Qingwen and Cai, Jisong and Chen, Li and Cui, Xiuqi and Ding, Yan and Feng, Siyuan and Gao, Shenyuan and He, Xindong and Hu, Xuan and Huang, Xu and others},
  journal={arXiv preprint arXiv:2503.06669},
  year={2025}
}

@article{bu2024robodual,
  title={Towards synergistic, generalized, and efficient dual-system for robotic manipulation},
  author={Bu, Qingwen and Li, Hongyang and Chen, Li and Cai, Jisong and Zeng, Jia and Cui, Heming and Yao, Maoqing and Qiao, Yu},
  journal={arXiv preprint arXiv:2410.08001},
  year={2024}
}

@article{bu2025univla,
  title={Univla: Learning to act anywhere with task-centric latent actions},
  author={Bu, Qingwen and Yang, Yanting and Cai, Jisong and Gao, Shenyuan and Ren, Guanghui and Yao, Maoqing and Luo, Ping and Li, Hongyang},
  journal={arXiv preprint arXiv:2505.06111},
  year={2025}
}

@article{chen2025robotwin2,
  title={Robotwin 2.0: A scalable data generator and benchmark with strong domain randomization for robust bimanual robotic manipulation},
  author={Chen, Tianxing and Chen, Zanxin and Chen, Baijun and Cai, Zijian and Liu, Yibin and Li, Zixuan and Liang, Qiwei and Lin, Xianliang and Ge, Yiheng and Gu, Zhenyu and others},
  journal={arXiv preprint arXiv:2506.18088},
  year={2025}
}

@article{zhong2025surveyvla,
  title={A survey on vision-language-action models: An action tokenization perspective},
  author={Zhong, Yifan and Bai, Fengshuo and Cai, Shaofei and Huang, Xuchuan and Chen, Zhang and Zhang, Xiaowei and Wang, Yuanfei and Guo, Shaoyang and Guan, Tianrui and Lui, Ka Nam and others},
  journal={arXiv preprint arXiv:2507.01925},
  year={2025}
}

@article{liu2025robovlms,
  title={What matters in building vision-language-action models for generalist robots},
  author={Li, Xinghang and Li, Peiyan and Qian, Long and Liu, Minghuan and Wang, Dong and Liu, Jirong and Kang, Bingyi and Ma, Xiao and Wang, Xinlong and Guo, Di and others},
  journal={Nature Machine Intelligence},
  pages={1--15},
  year={2026},
  publisher={Nature Publishing Group UK London}
}

@inproceedings{zhao2025cotvla,
  title={Cot-vla: Visual chain-of-thought reasoning for vision-language-action models},
  author={Zhao, Qingqing and Lu, Yao and Kim, Moo Jin and Fu, Zipeng and Zhang, Zhuoyang and Wu, Yecheng and Li, Zhaoshuo and Ma, Qianli and Han, Song and Finn, Chelsea and others},
  booktitle={Proceedings of the IEEE/CVF Conference on Computer Vision and Pattern Recognition (CVPR)},
  pages={1702--1713},
  year={2025}
}

@article{team2025geminirobotics,
  title={Gemini robotics: Bringing ai into the physical world},
  author={Team, Gemini Robotics and Abeyruwan, Saminda and Ainslie, Joshua and Alayrac, Jean-Baptiste and Arenas, Montserrat Gonzalez and Armstrong, Travis and Balakrishna, Ashwin and Baruch, Robert and Bauza, Maria and Blokzijl, Michiel and others},
  journal={arXiv preprint arXiv:2503.20020},
  year={2025}
}

@inproceedings{huangotter,
  title={OTTER: A vision-language-action model with text-aware visual feature extraction},
  author={Huang, Huang and Liu, Fangchen and Fu, Letian and Wu, Tingfan and Mukadam, Mustafa and Malik, Jitendra and Goldberg, Ken and Abbeel, Pieter},
  booktitle={International Conference on Machine Learning (ICML)}
}

@article{jiang2025rynnvla,
  title={RynnVLA-001: Using human demonstrations to improve robot manipulation},
  author={Jiang, Yuming and Huang, Siteng and Xue, Shengke and Zhao, Yaxi and Cen, Jun and Leng, Sicong and Li, Kehan and Guo, Jiayan and Wang, Kexiang and Chen, Mingxiu and others},
  journal={arXiv preprint arXiv:2509.15212},
  year={2025}
}

@article{generalist2025gen0,
          author = {Generalist AI Team},
          title = {GEN-0: Embodied foundation models that scale with physical interaction},
          journal = {Generalist AI Blog},
          year = {2025},
          note = {https://generalistai.com/blog/preview-uqlxvb-bb.html},
        }

@article{yang2025instructvla,
  title={Instructvla: Vision-language-action instruction tuning from understanding to manipulation},
  author={Yang, Shuai and Li, Hao and Chen, Yilun and Wang, Bin and Tian, Yang and Wang, Tai and Wang, Hanqing and Zhao, Feng and Liao, Yiyi and Pang, Jiangmiao},
  journal={arXiv preprint arXiv:2507.17520},
  year={2025}
}

@article{liu2025hybridvla,
  title={Hybridvla: Collaborative diffusion and autoregression in a unified vision-language-action model},
  author={Liu, Jiaming and Chen, Hao and An, Pengju and Liu, Zhuoyang and Zhang, Renrui and Gu, Chenyang and Li, Xiaoqi and Guo, Ziyu and Chen, Sixiang and Liu, Mengzhen and others},
  journal={arXiv preprint arXiv:2503.10631},
  year={2025}
}

@inproceedings{chen2025clutterdexgrasp,
  title        = {{ClutterDexGrasp}: A {Sim-to-Real} system for general dexterous grasping in cluttered scenes},
  author       = {Chen, Zeyuan and Yan, Qiyang and Chen, Yuanpei and Wu, Tianhao and Zhang, Jiyao and Ding, Zihan and Li, Jinzhou and Yang, Yaodong and Dong, Hao},
  booktitle    = {Conference on Robot Learning (CoRL)},
  pages        = {885--905},
  year         = {2025},
  organization = {PMLR}
}

@article{zhao2023act,
  title     = {Learning fine-grained bimanual manipulation with low-cost hardware},
  author    = {Zhao, Tony Z and Kumar, Vikash and Levine, Sergey and Finn, Chelsea},
  journal   = {Robotics: Science and Systems (RSS)},
  year      = {2023},
  publisher = {Robotics: Science and Systems Foundation}
}

@inproceedings{zhu2025scaledp,
  title={Scaling diffusion policy in transformer to 1 billion parameters for robotic manipulation},
  author={Zhu, Minjie and Zhu, Yichen and Li, Jinming and Wen, Junjie and Xu, Zhiyuan and Liu, Ning and Cheng, Ran and Shen, Chaomin and Peng, Yaxin and Feng, Feifei and others},
  booktitle={IEEE International Conference on Robotics and Automation (ICRA)},
  pages={10838--10845},
  year={2025},
  organization={IEEE}
}

@article{wu2025foresight,
  title={From foresight to forethought: Vlm-in-the-loop policy steering via latent alignment},
  author={Wu, Yilin and Tian, Ran and Swamy, Gokul and Bajcsy, Andrea},
  journal={arXiv preprint arXiv:2502.01828},
  year={2025}
}

@inproceedings{wanna2025lang,
  title={Let's talk about language! investigating linguistic diversity in embodied AI datasets},
  author={Wanna, Selma Liliane and Luhtaru, Agnes and Barron, Ryan and Salfity, Jonathan and Moore, Juston and Matuszek, Cynthia and Pryor, Mitch},
  booktitle={1st Workshop on Safely Leveraging Vision-Language Foundation Models in Robotics: Challenges and Opportunities}
}

@article{zhang2025align,
  title={Align-then-steer: Adapting the vision-language action models through unified latent guidance},
  author={Zhang, Yang and Wang, Chenwei and Lu, Ouyang and Zhao, Yuan and Ge, Yunfei and Sun, Zhenglong and Li, Xiu and Zhang, Chi and Bai, Chenjia and Li, Xuelong},
  journal={arXiv preprint arXiv:2509.02055},
  year={2025}
}

@inproceedings{nakamoto2024steering,
  title        = {Steering your generalists: Improving robotic foundation models via value guidance},
  author       = {Nakamoto, Mitsuhiko and Mees, Oier and Kumar, Aviral and Levine, Sergey},
  booktitle    = {Conference on Robot Learning (CoRL)},
  pages        = {4996--5013},
  year         = {2025},
  organization = {PMLR}
}

@article{geng2025roboverse,
  title={RoboVerse: Towards a unified platform, dataset and benchmark for scalable and generalizable robot learning},
  author={Geng, Haoran and Wang, Feishi and Wei, Songlin and Li, Yuyang and Wang, Bangjun and An, Boshi and Cheng, Charlie Tianyue and Lou, Haozhe and Li, Peihao and Wang, Yen-Jen and others},
  journal={arXiv preprint arXiv:2504.18904},
  year={2025}
}

@article{cen2025worldvla,
  title={WorldVLA: Towards autoregressive action world model},
  author={Cen, Jun and Yu, Chaohui and Yuan, Hangjie and Jiang, Yuming and Huang, Siteng and Guo, Jiayan and Li, Xin and Song, Yibing and Luo, Hao and Wang, Fan and others},
  journal={arXiv preprint arXiv:2506.21539},
  year={2025}
}

@inproceedings{zhang2025dreamvla,
  title     = {{DreamVLA}: A vision-language-action model dreamed with comprehensive world knowledge},
  author    = {Zhang, Wenyao and Liu, Hongsi and Qi, Zekun and Wang, Yunnan and Yu, Xinqiang and Zhang, Jiazhao and Dong, Runpei and He, Jiawei and Wang, He and Zhang, Zhizheng and others},
  booktitle = {Advances in Neural Information Processing Systems (NeurIPS)},
  year      = {2025}
}

@inproceedings{xu2021learning,
  title={Learning 3D dynamic scene representations for robot manipulation},
  author={Xu, Zhenjia and He, Zhanpeng and Wu, Jiajun and Song, Shuran},
  booktitle={Conference on Robot Learning (CoRL)},
  pages={126--142},
  year={2021},
  organization={PMLR}
}

@inproceedings{li2025cameras,
  title     = {Cameras as relative positional encoding},
  author    = {Li, Ruilong and Yi, Brent and Liu, Junchen and Gao, Hang and Ma, Yi and Kanazawa, Angjoo},
  booktitle = {Advances in Neural Information Processing Systems (NeurIPS)},
  year      = {2025}
}

@article{chen2025internvlam1,
  title={Internvla-m1: A spatially guided vision-language-action framework for generalist robot policy},
  author={Chen, Xinyi and Chen, Yilun and Fu, Yanwei and Gao, Ning and Jia, Jiaya and Jin, Weiyang and Li, Hao and Mu, Yao and Pang, Jiangmiao and Qiao, Yu and others},
  journal={arXiv preprint arXiv:2510.13778},
  year={2025}
}

@article{fei2025liberoplus,
  title={Libero-plus: In-depth robustness analysis of vision-language-action models},
  author={Fei, Senyu and Wang, Siyin and Shi, Junhao and Dai, Zihao and Cai, Jikun and Qian, Pengfang and Ji, Li and He, Xinzhe and Zhang, Shiduo and Fei, Zhaoye and others},
  journal={arXiv preprint arXiv:2510.13626},
  year={2025}
}

@article{qwen3vl2025,
  title={Qwen3-vl technical report},
  author={Bai, Shuai and Cai, Yuxuan and Chen, Ruizhe and Chen, Keqin and Chen, Xionghui and Cheng, Zesen and Deng, Lianghao and Ding, Wei and Gao, Chang and Ge, Chunjiang and others},
  journal={arXiv preprint arXiv:2511.21631},
  year={2025}
}

@article{sun2026vlajepa,
  title={{VLA-JEPA}: Enhancing vision-language-action model with latent world model},
  author={Sun, Jingwen and Zhang, Wenyao and Qi, Zekun and Ren, Shaojie and Liu, Zezhi and Zhu, Hanxin and Sun, Guangzhong and Jin, Xin and Chen, Zhibo},
  journal={arXiv preprint arXiv:2602.10098},
  year={2026}
}

@article{xu2025bayesvla,
  title={Seeing to act, prompting to specify: A Bayesian factorization of vision language action policy},
  author={Xu, Kechun and Zhu, Zhenjie and Chen, Anzhe and Zhao, Shuqi and Huang, Qing and Yang, Yifei and Lu, Haojian and Xiong, Rong and Tomizuka, Masayoshi and Wang, Yue},
  journal={arXiv preprint arXiv:2512.11218},
  year={2025}
}

@article{nasiriany2024robocasa,
  title={Robocasa: Large-scale simulation of everyday tasks for generalist robots},
  author={Nasiriany, Soroush and Maddukuri, Abhiram and Zhang, Lance and Parikh, Adeet and Lo, Aaron and Joshi, Abhishek and Mandlekar, Ajay and Zhu, Yuke},
  journal={arXiv preprint arXiv:2406.02523},
  year={2024}
}

@article{lian2026langforce,
  title   = {{LangForce}: {Bayesian} decomposition of vision language action
             models via latent action queries},
  author  = {Lian, Shijie and Yu, Bin and Lin, Xiaopeng and Yang, Laurence T.
             and Shen, Zhaolong and Wu, Changti and Miao, Yuzhuo and
             Huang, Cong and Chen, Kai},
  journal = {arXiv preprint arXiv:2601.15197},
  year    = {2026}
}

@article{orjuela2026brittle,
  title   = {Robust skills, brittle grounding: Diagnosing restricted
             generalization in vision-language action policies via
             multi-object picking},
  author  = {Orjuela, Santiago and others},
  journal = {arXiv preprint arXiv:2602.24143},
  year    = {2026}
}

@article{fang2026vision,
  title={When Vision Overrides Language: Evaluating and Mitigating Counterfactual Failures in VLAs},
  author={Fang, Yu and Feng, Yuchun and Jing, Dong and Liu, Jiaqi and Yang, Yue and Wei, Zhenyu and Szafir, Daniel and Ding, Mingyu},
  journal={arXiv preprint arXiv:2602.17659},
  year={2026}
}

@article{fu2026cap,
  title={CaP-X: A Framework for Benchmarking and Improving Coding Agents for Robot Manipulation},
  author={Fu, Max and Yu, Justin and El-Refai, Karim and Kou, Ethan and Xue, Haoru and Huang, Huang and Xiao, Wenli and Wang, Guanzhi and Li, Fei-Fei and Shi, Guanya and others},
  journal={arXiv preprint arXiv:2603.22435},
  year={2026}
}

@article{xu2021efficient,
  title={Efficient learning of goal-oriented push-grasping synergy in clutter},
  author={Xu, Kechun and Yu, Hongxiang and Lai, Qianen and Wang, Yue and Xiong, Rong},
  journal={IEEE Robotics and Automation Letters},
  volume={6},
  number={4},
  pages={6337--6344},
  year={2021},
  publisher={IEEE}
}

@article{intelligence2026pi07,
  title={$\pi_{0.7}$: a Steerable Generalist Robotic Foundation Model with Emergent Capabilities},
  author={Intelligence, Physical and Ai, Bo and Amin, Ali and Aniceto, Raichelle and Balakrishna, Ashwin and Balke, Greg and Black, Kevin and Bokinsky, George and Cao, Shihao and Charbonnier, Thomas and others},
  journal={arXiv preprint arXiv:2604.15483},
  year={2026}
}

@article{fang2025intention,
  title={From intention to execution: Probing the generalization boundaries of vision-language-action models},
  author={Fang, Irving and Zhang, Juexiao and Tong, Shengbang and Feng, Chen},
  journal={arXiv preprint arXiv:2506.09930},
  year={2025}
}

@article{guo2025robustness,
  title={On robustness of vision-language-action model against multi-modal perturbations},
  author={Guo, Jianing and Wu, Zhenhong and Tu, Chang and Ma, Yiyao and Kong, Xiangqi and Liu, Zhiqian and Ji, Jiaming and Zhang, Shuning and Chen, Yuanpei and Chen, Kai and others},
  journal={arXiv preprint arXiv:2510.00037},
  year={2025}
}

@article{zhan2026stable,
  title={Stable Language Guidance for Vision-Language-Action Models},
  author={Zhan, Zhihao and Chen, Yuhao and Zhou, Jiaying and Lv, Qinhan and Liu, Hao and Wang, Keze and Lin, Liang and Wang, Guangrun},
  journal={arXiv preprint arXiv:2601.04052},
  year={2026}
}

@article{wu2026pragmatic,
  title={A Pragmatic VLA Foundation Model},
  author={Wu, Wei and Lu, Fan and Wang, Yunnan and Yang, Shuai and Liu, Shi and Wang, Fangjing and Zhu, Qian and Sun, He and Wang, Yong and Ma, Shuailei and others},
  journal={arXiv preprint arXiv:2601.18692},
  year={2026}
}

@article{li2026causal,
  title={Causal World Modeling for Robot Control},
  author={Li, Lin and Zhang, Qihang and Luo, Yiming and Yang, Shuai and Wang, Ruilin and Han, Fei and Yu, Mingrui and Gao, Zelin and Xue, Nan and Zhu, Xing and others},
  journal={arXiv preprint arXiv:2601.21998},
  year={2026}
}

@inproceedings{songdenoising,
  title={Denoising Diffusion Implicit Models},
  author={Song, Jiaming and Meng, Chenlin and Ermon, Stefano},
  booktitle={International Conference on Learning Representations}
}

@inproceedings{tian2025internA1,
  title={Interndata-a1: Pioneering high-fidelity synthetic data for pre-training generalist policy},
  author={Tian, Yang and Yang, Yuyin and Xie, Yiman and Cai, Zetao and Shi, Xu and Gao, Ning and Liu, Hangxu and Jiang, Xuekun and Qiu, Zherui and Yuan, Feng and others},
  booktitle={Proceedings of the IEEE/CVF Conference on Computer Vision and Pattern Recognition},
  pages={976--985},
  year={2026}
}

@article{tan2025ript,
  title={Interactive post-training for vision-language-action models},
  author={Tan, Shuhan and Dou, Kairan and Zhao, Yue and Kr{\"a}henb{\"u}hl, Philipp},
  journal={arXiv preprint arXiv:2505.17016},
  year={2025}
}

@article{zheng2025xvla,
  title={X-vla: Soft-prompted transformer as scalable cross-embodiment vision-language-action model},
  author={Zheng, Jinliang and Li, Jianxiong and Wang, Zhihao and Liu, Dongxiu and Kang, Xirui and Feng, Yuchun and Zheng, Yinan and Zou, Jiayin and Chen, Yilun and Zeng, Jia and others},
  journal={arXiv preprint arXiv:2510.10274},
  year={2025}
}
